\documentclass[10pt,twocolumn,letterpaper]{article}

\usepackage[pagenumbers]{cvpr} 
\usepackage{multirow}
\usepackage{diagbox}
\usepackage{pifont}

\usepackage{float}
\usepackage[utf8]{inputenc}
\usepackage{etoolbox}
\usepackage{silence}
\makeatletter
\robustify\@latex@warning@no@line
\makeatother
\makeatletter
\def\thanks#1{\protected@xdef\@thanks{\@thanks
\protect\footnotetext{#1}}}
\makeatother
\usepackage{authblk}
\usepackage{color}
\definecolor{Green}{RGB}{0,176,80}
\definecolor{Purple}{RGB}{112,48,160}
\newcommand{\eat}[1]{}

\definecolor{cvprblue}{rgb}{0.21,0.49,0.74}
\definecolor{cvprred}{rgb}{0.894, 0.0, 0.498}
\usepackage[pagebackref,breaklinks,colorlinks,allcolors=cvprblue]{hyperref}

\def\paperID{11227} 
\def\confName{CVPR}
\def\confYear{2025}

\title{Omni-Scene: Omni-Gaussian Representation for \\Ego-Centric Sparse-View Scene Reconstruction}

\author[1,3]{Dongxu Wei}
\author[1,2]{Zhiqi Li}
\author[1*]{Peidong Liu\thanks{*Corresponding author}}

\affil[1]{School of Engineering, Westlake University}
\affil[2]{College of Computer Science and Technology, Zhejiang University}
\affil[3]{Institute of Advanced Technology, Westlake Institute for Advanced Study \authorcr
  \{\tt weidongxu, lizhiqi49, liupeidong\}@westlake.edu.cn \authorcr \href{https://wswdx.github.io/omniscene}{\textcolor{cvprred}{https://wswdx.github.io/omniscene}}}

\let\oldtwocolumn\twocolumn
\renewcommand\twocolumn[1][]{%
    \oldtwocolumn[{#1}{
    \begin{center}
           \vspace{-27pt}
           \includegraphics[width=0.99\textwidth]{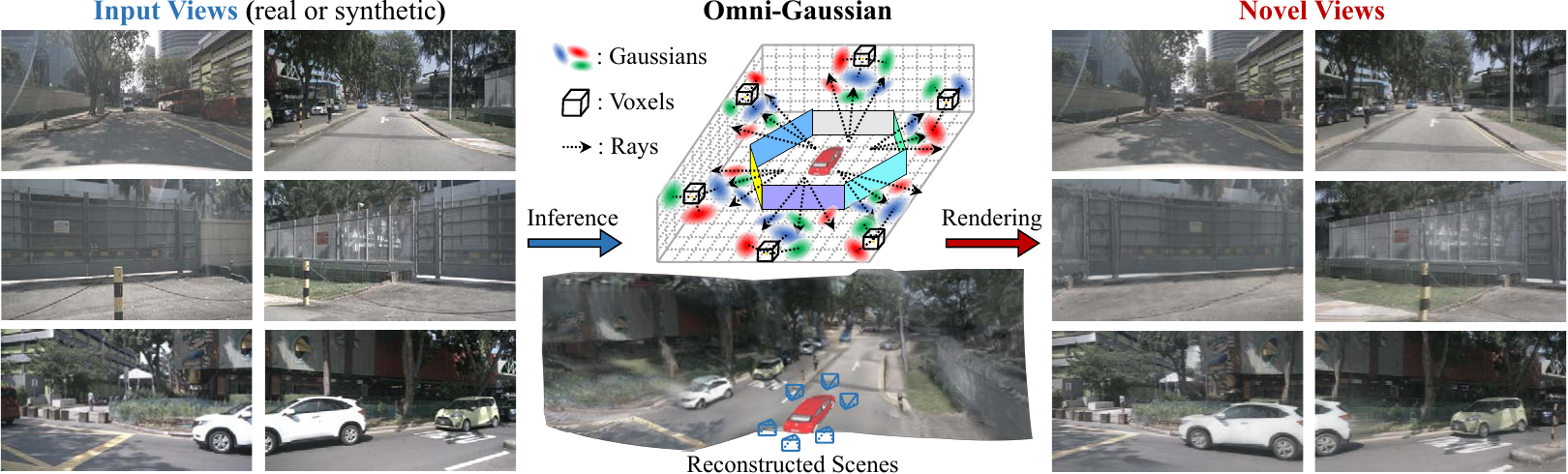}
           \vspace{-5pt}
           \captionof{figure}{Provided with six surrounding images captured in real world or synthesized by 2D diffusion models, we can generate high-quality 3D Gaussians based on our Omni-Gaussian representation for ego-centric scene reconstruction and novel view synthesis.}
           \label{Fig.fig0}
        \end{center}
    }]
}
\begin{document}

\maketitle

\begin{abstract}
Prior works employing pixel-based Gaussian representation have demonstrated efficacy in feed-forward sparse-view reconstruction. However, such representation necessitates cross-view overlap for accurate depth estimation, and is challenged by object occlusions and frustum truncations. As a result, these methods require scene-centric data acquisition to maintain cross-view overlap and complete scene visibility to circumvent occlusions and truncations, which limits their applicability to scene-centric reconstruction.
In contrast, in autonomous driving scenarios, a more practical paradigm is ego-centric reconstruction, which is characterized by minimal cross-view overlap and frequent occlusions and truncations.
The limitations of pixel-based representation thus hinder the utility of prior works in this task.
In light of this, this paper conducts an in-depth analysis of different representations, and introduces Omni-Gaussian representation with tailored network design to complement their strengths and mitigate their drawbacks.
Experiments show that our method significantly surpasses state-of-the-art methods, pixelSplat and MVSplat, in ego-centric reconstruction, and achieves comparable performance to prior works in scene-centric reconstruction.
\end{abstract}

\vspace{-10pt}
\section{Introduction}
\label{sec:intro}
Reconstructing 3D scenes from sparse observations is a crucial task in computer vision and graphics.
Recent efforts \cite{pixelnerf2021,ibrnet2021,nrays2022,mvsnerf2021,geonerf2022,efficient2022,nerfusion2022,lrm2023,murf2024,lf2022,gpnr2022,attnrend2023,pixelsplat2024,mvsplat2024,flash3d2024,lgm2024,splatterimage2024,grm2024,geolrm2024,tmgs2024,mvsgs2024,ggrt2024} have integrated 3D structural priors as inductive biases into neural networks, enabling the prediction of implicit neural field \cite{nerf2021}, light field \cite{lf2022}, or explicit 3D Gaussians \cite{3dgs2023} for scene reconstruction in a single forward pass.
\begin{figure*}[!ht]
\centering
\vspace{-15pt}
\includegraphics[width=0.99\textwidth]{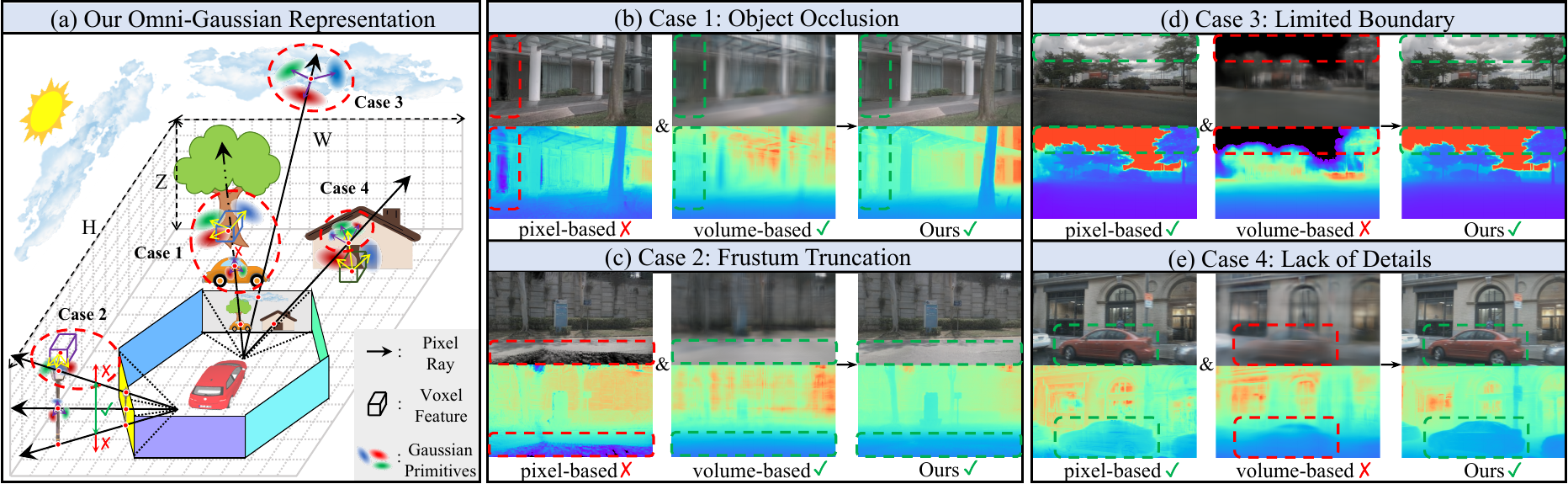}
\caption{Illustration of our Omni-Gaussian representation. Our Omni-Gaussian incorporates two representations, pixel-based and volume-based Gaussians. In (a), we illustrate bad cases when relying solely on one representation (i.e., Case 1 and 2 for pixel-based, Case 3 and 4 for volume-based), and how we use the other one to compensate for the shortcomings. In (b)-(e), we present examples of these cases under the task of ego-centric driving scene reconstruction. \emph{Green dashed lines} denote areas plausibly rendered in novel views, while \emph{red ones} highlight undesirable artifacts due to weaknesses of pixel-based or volume-based Gaussian. We can observe that Omni-Gaussian leveraging the complementary nature of the two representations can achieve optimal results for all cases.}
\label{Fig.fig1}
\vspace{-10pt}
\end{figure*}
Notably, due to the efficiency of rasterization-based rendering and the explicit nature of 3D Gaussians \cite{3dgs2023}, Gaussian-based methods \cite{pixelsplat2024,mvsplat2024,flash3d2024,lgm2024,splatterimage2024,grm2024,geolrm2024,tmgs2024,mvsgs2024,ggrt2024} have shown superiority in both inference speed and visual quality compared to those based on neural field \cite{pixelnerf2021,ibrnet2021,nrays2022,mvsnerf2021,geonerf2022,efficient2022,nerfusion2022,lrm2023,murf2024} or light field \cite{lf2022,gpnr2022,attnrend2023}.
Typically, these methods assume existence of large overlaps among the observed input views. Thus they can utilize techniques such as multi-view cross attention \cite{lgm2024,grm2024,ggrt2024}, epipolar lines \cite{pixelsplat2024} or cost volumes \cite{mvsplat2024,mvsgs2024} to learn pixel-level cross-view correlation, and then infer per-pixel depths with proper scales. Hence they can further predict per-pixel Gaussians and use depths to unproject them to 3D along pixel rays for scene reconstruction.
A common feature for all of these methods is the use of pixel-based Gaussian representation.

Although works utilizing the pixel-based Gaussian representation have achieved great success, they pose strong hypothesis regarding the existence of large overlaps among input views. This implies the necessity of capturing input-view images encircling the scene. Otherwise, they would fail to predict accurate per-pixel depths due to the scale ambiguity \cite{pixelsplat2024}.
In contrast to such scene-centric reconstruction, a more practical case especially for autonomous driving systems, is ego-centric reconstruction, where we can only acquire input views from cameras rigidly mounted around the car, with minimal overlaps ({\textless}15\%) existing only between adjacent cameras.
As evident in Sec.\ref{sec:experiment:main}, previous methods with pixel-based representation would fail for ego-centric reconstruction.
Despite the difficulty in predicting per-pixel depths, their failure can be attributed to two underlying weaknesses inherent in the pixel-based representation as showcased by Case 1 and 2 in Fig.\ref{Fig.fig1}.
In Case 1, when object in the target novel view is occluded in the input view (e.g., tree behind the car in Fig.\ref{Fig.fig1}(a)), pixel-based representation can only rely on 2D local features of the non-occluded object for inferring the occluded one, which fails especially when their appearances are far different from each other.
In Case 2, when object in the novel view falls outside of the input view frustum (e.g., top of the streetlight in Fig.\ref{Fig.fig1}(a)), pixel-based representation cannot predict positions of Gaussians through unprojection along pixel rays.

These two cases also pose challenges for 3D perception tasks like 3D object detection \cite{bevformer2022,bevfusion2023,petr2022} and occupancy prediction \cite{surroundocc2023,occformer2023,tpvformer2023}, which require to perceive partially occluded or truncated objects.
Existing 3D perception works resort to volume-based representations like bird's eye view (BEV) grids \cite{bevformer2022,bevfusion2023,petr2022} and 3D voxels \cite{surroundocc2023,occformer2023,tpvformer2023} as the solution.
Since volume is spatially-continuous in 3D space, contents absent in the 2D inputs can be supplemented by their neighbors at the 3D level.
Besides, with camera projection knowledge to enable 3D-to-2D cross attention \cite{bevformer2022}, we can directly lift 2D features to 3D space instead of relying on cross-view overlap for depth-based 2D-to-3D unprojection.
Intuitively, we conjecture that we can utilize a volume-based Gaussian representation in the reconstruction task (i.e., represent Gaussians with voxels in the volume) to minimize dependence on cross-view overlap and mitigate bad effects brought by occlusions and truncations.
However, as illustrated by Case 3 and 4 in Fig.\ref{Fig.fig1}, this representation also has drawbacks. Due to the bounded nature of volume (i.e., bounded within the range of $H{\times}W{\times}Z$ around the car), volume-based Gaussian cannot reconstruct elements far away from the car (e.g., sky in Case 3 of Fig.\ref{Fig.fig1}(a)).
Besides, encoding features for a volume with cubic complexity limits the volume resolution, potentially resulting in the lack of details (e.g., house in Case 4 of Fig.\ref{Fig.fig1}(a)).

In this paper, considering limitations of pixel and volume-based Gaussians, we propose Omni-Scene, which employs Omni-Gaussian representation and tailored network designs to reach the best of both worlds.
\emph{The core lies in how to optimize volume and pixel-based Gaussians to their full potential, and leverage their unique attributes to enable their collaboration.}
Specifically, for volume-based Gaussian, we propose Volume Builder composed of Triplane Transformer and Volume Decoder to reconstruct coarse 3D structures with voxel-anchored Gaussians.
In particular, our Triplane Transformer uses triplane as a light-weight alternative of volume, where we employ cross-image and cross-plane deformable attentions to enhance volumetric feature encoding.
For pixel-based Gaussian, we propose Pixel Decorator, which complements volume-based Gaussian with distant elements and better details. Our Pixel Decorator comprises Multi-View U-Net and Pixel Decoder, responsible for cross-view attended feature extraction and per-pixel Gaussian prediction, respectively.
To enable collaboration between the two representations, we introduce Projection-Based Feature Fusion and Depth-Guided Training Decomposition for their seamless fusion and better complementarity, thereby boosting the performance.
In summary, our main contributions are as follows:
\begin{itemize}
\item We propose Omni-Scene, an Omni-Gaussian representation with tailored network design for ego-centric reconstruction, taking advantages of both pixel and volume-based representations while eliminating their drawbacks.
\item We introduce a novel ego-centric reconstruction task to a popular driving dataset (i.e., nuScenes \cite{nusc2020}), with the aim of scene-level 3D reconstruction and novel view synthesis given only single-frame surrounding images.
We hope this can facilitate further research in this field.
\item Experiments show that our method significantly outperforms state-of-the-art feed-forward reconstruction methods including pixelSplat \cite{pixelsplat2024} and MVSplat \cite{mvsplat2024} on the ego-centric task.
We also achieve competitive performance with prior works on the scene-centric task performed on RealEstate10K dataset \cite{re10k2018}.
\end{itemize}  
\section{Related Work}
\label{sec:relatedwork}

\noindent\textbf{Neural Reconstruction and Rendering.}
Recent approaches \cite{nerf2021,mipnerf2021,zipnerf2023,instantngp2022,3dgs2023,mipsplat2024,2dgs2024} leveraging neural rendering and reconstruction techniques can model scenes as learnable 3D representations, and achieve 3D reconstruction and novel view synthesis through iterative back propagation.
NeRF \cite{nerf2021} has been recognized for its ability to capture high-frequency details in reconstructed scenes.
However, it requires dense queries for each ray during rendering, which, despite subsequent efforts for acceleration \cite{zipnerf2023,instantngp2022}, still results in high computational demand that limits its real-time capability.
3D Gaussian Splatting (3DGS) \cite{3dgs2023} mitigates this issue by explicitly modeling scenes with 3D Gaussians and employing an efficient rasterization-based rendering pipeline.
Although 3DGS and NeRF, along with their variants \cite{mipnerf2021,mipsplat2024,emernerf2023,drivegs2024}, have demonstrated superior performance in single-scene reconstruction, they usually require per-scene optimization and dense scene captures, making the reconstruction process time-consuming and unscalable.
Different from these works, our method can reconstruct 3D scenes from sparse observations in a single forward pass.

\noindent\textbf{Feed-Forward Reconstruction with Implicit 3D Representations.}
This line of works incorporate implicit 3D priors, such as NeRF \cite{nerf2021} or light field \cite{lf2022}, into their networks to achieve feed-forward reconstruction.
NeRF-based methods \cite{pixelnerf2021,ibrnet2021,nrays2022,mvsnerf2021,geonerf2022,efficient2022,nerfusion2022,lrm2023,murf2024} leverage transformers with multi-view cross attentions \cite{lrm2023,lgm2024,grm2024,ggrt2024}, or employ projective 3D priors like epipolar lines \cite{pixelnerf2021,ibrnet2021,nrays2022,pixelsplat2024} and cost volumes \cite{mvsnerf2021,geonerf2022,efficient2022,nerfusion2022,mvsplat2024,mvsgs2024} to estimate radiance fields for reconstruction, which inherits the expensive ray querying process of NeRF rendering.
Consequently, these methods are exceedingly time-consuming during both training and inference phases.
In contrast, light field-based approaches \cite{lf2022,gpnr2022,attnrend2023} can bypass NeRF rendering by directly regressing per-ray colors based on ray-to-image cross attentions, which sacrifices interpretability for efficiency.
However, due to the lack of interpretable 3D structure, they fail to reconstruct 3D geometries of scenes.

\noindent\textbf{Feed-Forward Reconstruction with 3D Gaussians.}
Recent methods \cite{pixelsplat2024,mvsplat2024,flash3d2024,lgm2024,splatterimage2024,grm2024,geolrm2024,tmgs2024,mvsgs2024,ggrt2024} utilizing 3DGS can achieve both interpretability and efficiency. Typically, they adopt 3D priors akin to NeRF-based methods (e.g., epipolar lines \cite{pixelsplat2024}, cost volumes \cite{mvsplat2024,mvsgs2024} and multi-view cross attentions \cite{lgm2024,grm2024,ggrt2024}) into their networks, and employ pixel-based Gaussian representation to predict per-pixel Gaussians along the rays for reconstruction.
However, such pixel-based representation depends on large cross-view overlap to predict depths, and suffers from object occlusion and frustum truncation, thus only suits for scene-centric reconstruction with limited applicability.
In contrast, this paper concentrates on ego-centric reconstruction, which is characterized by minimal cross-view overlap and frequent occurrences of object occlusion and frustum truncation.
This has motivated our research into a novel 3D representation that is not overly dependent on cross-view overlap, and can address the limitations of pixel-based representation in the meantime.

\noindent\textbf{Neural Representation in 3D Perception.}
Similar to 3D reconstruction from multi-view images, 3D perception works \cite{lss2020,petr2022,bevformer2022,bevfusion2023,tpvformer2023,occformer2023,surroundocc2023} also utilize multi-view images as input and perform 3D perception tasks like 3D detection \cite{petr2022,bevformer2022,bevfusion2023}, map segmentation \cite{lss2020,bevformer2022,bevfusion2023}, and 3D occupancy prediction \cite{tpvformer2023,occformer2023,surroundocc2023}.
Early 3D perception attempts \cite{lss2020,petr2022,bevfusion2023} like Lift-Splat-Shoot (LSS)\cite{lss2020} employ pixel-wise depths to unproject pixel-wise features to 3D along camera rays, and project them onto BEV plane to enable 3D-level estimation. Similar to pixel-based representation in 3D reconstruction, such pixel-based approach would fail in cases of object occlusion.
Recent 3D perception methods \cite{bevformer2022,tpvformer2023,occformer2023,surroundocc2023} manage to bypass pixel-wise unprojection that sensitive to occlusion. In particular, they directly encode feature at 3D level by employ volume-based representation (e.g., BEV grids \cite{bevformer2022} or 3D voxels \cite{tpvformer2023,occformer2023,surroundocc2023}), and achieves better performance especially when some of the objects are occluded by those closer to the camera.
Although these methods show potential to accurate 3D perception, the perception task itself is much more coarse-grained compared to 3D reconstruction task, making low-resolution volume sufficient for perception-oriented feature modeling.
\begin{figure*}[!ht]
\centering
\vspace{-18pt}
\includegraphics[width=0.99\textwidth]{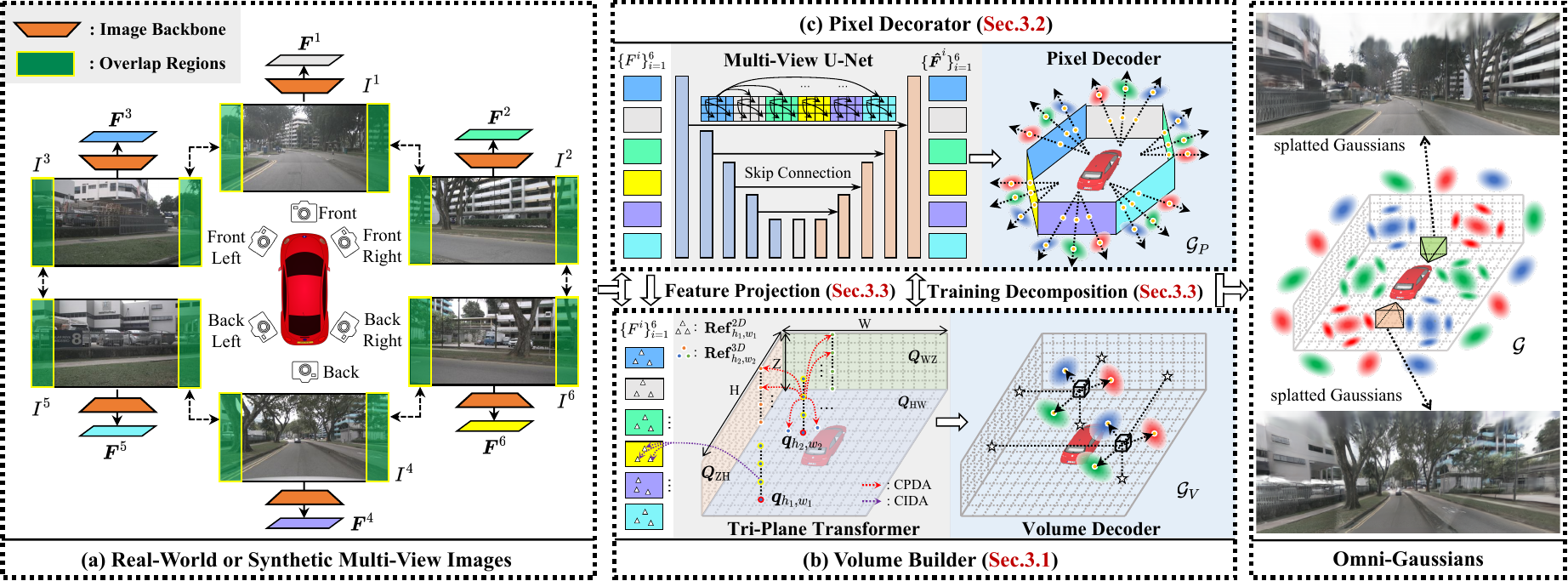}
\vspace{-3pt}
\caption{Overview. (a) Obtain images $\{\boldsymbol{I}^i\}^K_{i=1}$ from surrounding cameras with minimal overlap (e.g., adjacent image areas enclosed by \emph{green rectangles}) in a single frame, and extract 2D features using image backbone. (b) For Volume Builder, we first use Triplane Transformer to lift 2D features $\{\boldsymbol{F}^i\}^K_{i=1}$ to 3D volume space compressed by three orthogonal planes, where we employ cross-image and cross-plane deformable attentions to enhance feature encoding. Then, Volume Decoder takes voxels as anchors, and predict nearby Gaussians $\mathcal{G}_V$ for each voxel given features sampled from the three planes through bilinear interpolation. (c) For Pixel Decorator, we use Multi-View U-Net to propagate information across views and extract multiple 2D features for Pixel Decoder to predict pixel-based Gaussians $\mathcal{G}_P$ along rays. Through Volume-Pixel Collaborations including Projection-Based Feature Fusion and Depth-Guided Training Decomposition, we can make $\mathcal{G}_V$ and $\mathcal{G}_P$ complement for each other, and obtain the full Omni-Gaussians $\mathcal{G}$ for novel-view rendering.}
\label{Fig.fig2}
\vspace{-10pt}
\end{figure*}
In contrast, this paper focuses on 3D reconstruction task that requires fine-grained feature modeling, which exceeds the capability of volume representation.
\section{Method}
\label{sec:method}
\textbf{The overall pipeline} of Omni-Scene, a feed-forward approach to ego-centric sparse-view reconstruction, is shown in Fig.\ref{Fig.fig2}.
As depicted in Fig.\ref{Fig.fig2}(a), we accept $K$ surrounding images $\mathcal{I}=\{\boldsymbol{I}^i\}^K_{i=1}$ as inputs, which are captured or synthesized within a single frame.
We utilize a ResNet-50 \cite{resnet2016} backbone pre-trained with DINO objective \cite{dino2021} to extract 4$\times$ downsampled features $\mathcal{F}=\{\boldsymbol{F}^i\}^K_{i=1}$ for $\mathcal{I}$.
Then, as detailed in Fig.\ref{Fig.fig2}(b)-(c), the features are shared and fed into our Volume Builder (Sec.\ref{sec:method:vol}) and Pixel Decorater (Sec.\ref{sec:method:pix}) to predict volume-based Gaussians $\mathcal{G}_V$ and pixel-based Gaussians $\mathcal{G}_P$, respectively.
Utilizing Volume-Pixel Collaboration designs (Sec.\ref{sec:method:col}) including Projection-Based Feature Fusion and Depth-Guided Training Decomposition, we enable feature interaction between $\mathcal{G}_V$ and $\mathcal{G}_P$, and distinguish their attributes during training.
By fusing $\mathcal{G}_V$ and $\mathcal{G}_P$, we can obtain Omni-Gaussians $\mathcal{G}$ for reconstruction.
\subsection{Volume Builder}
\label{sec:method:vol}
Our Volume Builder aims to predict coarse 3D structures with volume-based Gaussians.
The primary challenge is how to lift 2D multi-view image features to the 3D volume space without explicitly maintaining dense voxels. We address this using Triplane Transformer.
Then, Volume Decoder is proposed to predict voxel-anchored Gaussians $\mathcal{G}_V$.

\noindent\textbf{Triplane Transformer.}
Representing volume as voxels and encoding features for each is expensive due to the cubic complexity of $H{\times}W{\times}Z$.
Therefore, we resort to triplane to disentangle volume into three axis-aligned orthogonal planes $HW$, $ZH$ and $WZ$.
Some object-level 3D reconstruction works \cite{lrm2023,dmv3d2024,crm2024} also adopt triplane representation to compress volume.
However, they either rely on dense per-pixel cross attention between triplanes and images \cite{dmv3d2024,crm2024}, or require input images to be also axis-aligned with triplanes \cite{lrm2023} for direct 2D-level feature encoding.
Neither of them is suitable for real-world scenes with much larger volume scales and unconstrained data collection.

Inspired by recent 3D perception methods \cite{bevformer2022,tpvformer2023} that replace global full-image attention with local deformable attention to efficiently lift information from 2D to 3D, our Triplane Transformer also utilize deformable attention to enable sparse but effective spatial correlations between 2D and 3D spaces.
Here we take the feature encoding of $HW$ plane as an example for explanation.
As shown in Fig.\ref{Fig.fig2}(b), we define a group of grid-shaped learnable embeddings $\boldsymbol{Q}_{HW}\in\mathbb{R}^{H{\times}W{\times}C}$ as the plane queries of transformer, where $C$ denotes the embedding channels.
Then, for query $\boldsymbol{q}_{h,w}$ positioned at $(h, w)$, we expand it to multiple 3D pillar points evenly spread along the $Z$ axis, and calculate their reference points $\boldsymbol{\rm{Ref}}^{2D}_{h,w}$ in 2D space by projecting them back to the input views.
Due to the sparse nature of such perspective projection, only the most relevant 2D features from 1$\sim$2 input views will be attended for $\boldsymbol{q}_{h,w}$, balancing efficiency and feature expressiveness.
The above operation, namely Cross-Image Deformable Attention (CIDA), is denoted by \emph{purple dashed arrows} in Fig.\ref{Fig.fig2}(b). We derive it as follows:
\begin{small}
\begin{equation}
  \boldsymbol{q}^{CIDA}_{h,w} = \frac{1}{K'}{\sum\limits^{K'}_{i=1}}\rm{DA}(\boldsymbol{q}_{h,w},\boldsymbol{\rm{Ref}}^{2D}_{h,w,i},\boldsymbol{F}_i),
  \label{eq:attn_img}
\end{equation}
\end{small}where $K'$, $\boldsymbol{\rm{Ref}}^{2D}_{h,w,i}$, $\rm{DA}$ represent the number of correlated views, 2D reference points in the $i$-th correlated view and deformable attention function, respectively.

Considering query pillar points might be occluded or located beyond the frustum range for any of the input views, we further utilize Cross-Plane Deformable Attention (CPDA) to enrich these points with cross-plane context.
In particular, for query $\boldsymbol{q}_{h,w}$, we project its coordinate $(h,w)$ onto the $HW$, $ZH$ and $WZ$ planes to obtain three sets of reference points $\boldsymbol{\rm{Ref}}^{3D}_{h,w}=\boldsymbol{\rm{Ref}}^{HW}_{h,w}\;{\cup}\;\boldsymbol{\rm{Ref}}^{ZH}_{h,w}\;{\cup}\;\boldsymbol{\rm{Ref}}^{WZ}_{h,w}$.
Here, $\boldsymbol{\rm{Ref}}^{HW}_{h,w}$ denotes neighbors of $\boldsymbol{q}_{h,w}$ within the $HW$ plane. $\boldsymbol{\rm{Ref}}^{ZH}_{h,w}$ and $\boldsymbol{\rm{Ref}}^{WZ}_{h,w}$ are orthogonal projections onto the $ZH$ and $WZ$ planes, derived from pillar points of $(h,w)$ evenly sampled along the $Z$ axis.
Utilizing $\boldsymbol{\rm{Ref}}^{3D}_{h,w}$, we extract contextual information from different planes, thereby enhancing the features as denoted by \emph{red dashed arrows} in Fig.\ref{Fig.fig2}(b). We derive it as follows:
\begin{small}
\begin{equation}
  \boldsymbol{q}^{CPDA}_{h,w} =\rm{DA}(\boldsymbol{q}_{h,w},\boldsymbol{\rm{Ref}}^{3D}_{h,w},\boldsymbol{Q}_{HW},\boldsymbol{Q}_{ZH},\boldsymbol{Q}_{WZ}),
  \label{eq:attn_plane}
\end{equation}
\end{small}where $\boldsymbol{Q}_{ZH},\boldsymbol{Q}_{WZ}$ denote queries of the other two planes.

Repeating these two cross attentions for queries of all the planes, we can obtain triplane feature with rich semantic and spatial context without dependency on cross-view overlap, which is necessary for previous approaches \cite{pixelsplat2024,mvsplat2024} that solely relied on pixel-based Gaussian representation.

\noindent\textbf{Volume Decoder.}
Our Volume Decoder is then proposed to estimate voxel-anchored Gaussians.
Specifically, given a voxel located at $(h,w,z)$, we first project its coordinate onto the three planes to obtain plane features through bilinear interpolation, which is followed by plane-wise summation to derive the aggregated voxel feature $\boldsymbol{f}_{h,w,z}$.
Then, we append three linear layers to $\boldsymbol{f}_{h,w,z}$ to predict parameters $(\boldsymbol{\delta}_v,\boldsymbol{\alpha}_v,\boldsymbol{s}_v,\boldsymbol{q}_v,\boldsymbol{c}_v)\}^V_{v=1}$ for $V$ Gaussians $\{\boldsymbol{G}^v\}^V_{v=1}$.
Each gaussian $\boldsymbol{G}^v$ is anchored near $(h,w,z)$ and shifted to a new position $\boldsymbol{\mu}_v$ according to the offset $\boldsymbol{\delta}_v\in\mathbb{R}^3$. The remaining parameters $\boldsymbol{\alpha}_v$, $\boldsymbol{s}_v$, $\boldsymbol{q}_v$, $\boldsymbol{c}_v$ denote opacity, scale, rotation quaternion and RGB color, respectively.
The same operation is repeated for all the voxels to obtain our volume-based Gaussians $\mathcal{G}_V\in\mathbb{R}^{H{\times}W{\times}Z\times{V}\times{D}}$, where $D$ is the dimension of Gaussian paramters.

\subsection{Pixel Decorator}
\label{sec:method:pix}
Our pixel decorator consists of Multi-View U-Net and Pixel Decoder, responsible for extracting cross-view correlated features and predicting pixel-based Gaussians $\mathcal{G}_P$, respectively.
Since $\mathcal{G}_P$ is obtained in alignment with fine-grained image space, it can add details to coarse voxel-anchored Gaussians $\mathcal{G}_V$.
Besides, since $\mathcal{G}_P$ can be unprojected to positions at infinite distance, it can supplement volume-bounded $\mathcal{G}_V$ with distant Gaussians.

\noindent\textbf{Multi-View U-Net.}
The Multi-View U-Net concatenates image features $\{\boldsymbol{F}^i\}^K_{i=1}$ and Pl$\ddot{\rm{u}}$cker ray embeddings $\{\boldsymbol{S}^i\}^K_{i=1}$ as inputs, where $\{\boldsymbol{S}^i\}^K_{i=1}$ can provide additional camera pose information \cite{lgm2024}.
Inspired by the patchified token compression introduced by a recent 2D diffusion transformer method \cite{pixart2024}, we apply patchified cross attentions to our Multi-View U-Net for efficient cross-view correlation as shown in Fig.\ref{Fig.fig2}(c).
Then, we can obtain 3D-aware features $\{\boldsymbol{\hat{F}}^i\}^K_{i=1}$ for each input view to decode Gaussians.

\noindent\textbf{Pixel Decoder.}
Our Pixel Decoder first upsamples the U-Net features $\{\boldsymbol{\hat{F}}^i\}^K_{i=1}$ to the original image resolution through bilinear interpolation, followed by several convolution layers to decode per-pixel depth $\boldsymbol{d}_p$ and Gaussian parameters $(\boldsymbol{\delta}_p, \boldsymbol{\alpha}_p, \boldsymbol{s}_p, \boldsymbol{q}_p, \boldsymbol{c}_p)$ for each Gaussian $\boldsymbol{G}^p$.
To obtain the center position $\boldsymbol{\mu}_p$, we first use $\boldsymbol{d}_p$ to unproject the pixel from the ray origin $\boldsymbol{o}_p$ to a coarse position along the ray direction $\boldsymbol{r}_p$, and then refine it with the learned offset $\boldsymbol{\delta}_p{\in}\mathbb{R}^3$. The unprojection process is derived as follows:
\begin{small}
\begin{equation}
  \boldsymbol{\mu}_p = \boldsymbol{o}_p + \boldsymbol{d}_p{\cdot}\boldsymbol{r}_p + \boldsymbol{\delta}_p.
  \label{eq:unproject}
\end{equation}
\end{small}Moreover, compared to predicting $\boldsymbol{d}_p$ from scratch, we find replacing it with the noisy estimation of a 2D foundation model \cite{metric3d2024} is beneficial for the performance, demonstrating the importance of Gaussian initialization \cite{3dgs2023}.
By doing the same to pixels of all the input views, we can obtain the pixel-based Gaussians $\mathcal{G}_P\in\mathbb{R}^{K{\times}R{\times}D}$, where $R$ is the total number of rays in an input view.

\subsection{Volume-Pixel Collaboration}
\label{sec:method:col}
The core of Omni-Gaussian representation lies in the collaboration of volume and pixel-based Gaussian representations.
For this purpose, we propose a dual approach that enables the collaboration from two aspects: Projection-Based Feature Fusion and Depth-Guided Training Decomposition.

\noindent\textbf{Projection-Based Feature Fusion.}
Our Volume Builder is expected to predict Gaussians at positions occluded or truncated in input views, which exceeds the design purpose of Pixel Decorator.
Therefore, to make Volume Builder aware of where the occlusion or truncation occurs, we propose to fuse the triplane queries $\boldsymbol{Q}_{HW},\boldsymbol{Q}_{ZH},\boldsymbol{Q}_{WZ}$ with projected features of pixel-based Gaussians $\mathcal{G}_P$.
Taking the plane of $HW$ as an example, we first filter out Gaussians fallen beyond the volume range of $H{\times}W{\times}Z$ for $\mathcal{G}_P$. Then, we collect U-Net features for the remaining Gaussians of $\mathcal{G}_P$ and project them onto the $HW$ plane. Features projected to the same query positions are averaged pooled and added to the corresponding query of $\boldsymbol{Q}_{HW}$ after a linear layer transformation.
The same process is applied to the $ZH$ and $WZ$ planes.
We demonstrate in our experiments (Sec.\ref{sec:experiment:abl}) that such feature fusion facilitates a complementary interaction between $\mathcal{G}_V$ and $\mathcal{G}_P$, thereby enhancing performance.

\noindent\textbf{Depth-Guided Training Decomposition.}
To further strengthen the collaboration, we propose a Depth-Guided Training Decomposition method to decompose our training objective based on the distinct spatial attributes of pixel and volume-based Gaussians.
\begin{figure*}[ht]
\centering
\vspace{-15pt}
\includegraphics[width=0.99\textwidth]{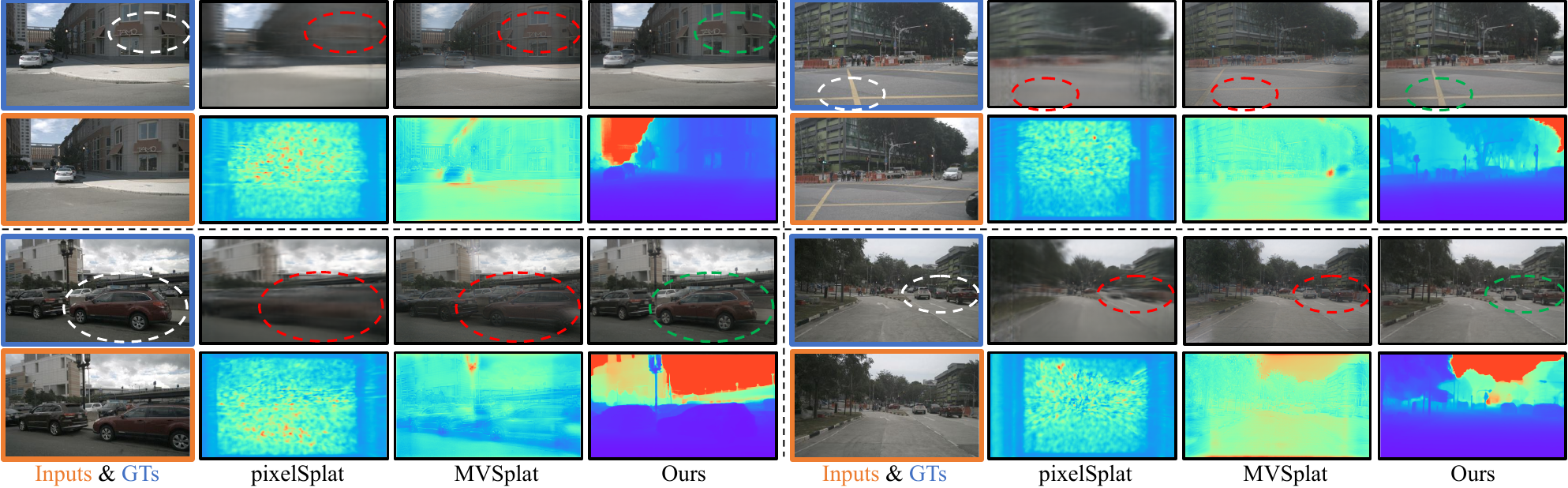}
\vspace{-5pt}
\caption{Comparisons on nuScenes \cite{nusc2020}. Images of input views (Inputs) and ground-truth novel views (GTs) are outlined by orange and blue rectangles, respectively. The remaining are generated novel views and depth maps (warmer colors denote greater distance while the opposite for cooler colors). The \emph{red dashed circles} denote undesirable artifacts, while the \emph{green ones} denote plausibly-rendered areas.}
\label{Fig.fig3}
\vspace{-12pt}
\end{figure*}
Specifically, due to the limited volume range, $\mathcal{G}_V$ is not supposed to reconstruct distant elements, which should be supplemented by $\mathcal{G}_P$ that has no distance limitation.
To achieve that, we first use $\mathcal{G}_P$ to render depth maps $\boldsymbol{\hat{D}}=\{\boldsymbol{\hat{D}}^i\}^{K}_{i=1}$ for all of the $K$ input views.
Then, we obtain 3D positions for all the pixels by unprojecting them along ray directions according to the estimated depths.
By assigning pixels located within the range of $H{\times}W{\times}Z$ to $1$, and assigning the remaining pixels to $0$, we can obtain masks $\boldsymbol{\hat{M}}=\{\boldsymbol{\hat{M}}^i\}^{K}_{i=1}$.
To enable appropriate supervision for volume-based Gaussian, we use $\boldsymbol{\hat{M}}$ to calculate masked photometric losses (i.e., mean squared error $L^{mse}_V$ and LPIPS loss $L^{lpips}_V$ \cite{lpips2018}) as well as masked L1 depth loss $L^{dpt}_V$ for input-view images and depths rendered from $\mathcal{G}_V$, where only pixels with mask values equal to $1$ will be used for loss calculation.
Note that $L^{dpt}_V$ denotes L1 errors between depths independently rendered from $\mathcal{G}_P$ and $\mathcal{G}_V$, which aims to align $\mathcal{G}_P$ and $\mathcal{G}_V$ to the same scale, and requires no external depth signals for supervision.
Combining with the photometric losses $L^{mse}_{full}$ and $L^{lpips}_{full}$ for novel-view images rendered from our full Gaussians $\mathcal{G}=\mathcal{G}_V \;{\cup}\;\mathcal{G}_P$, the overall training objective $L$ can be derived as follows:
\begin{small}
\begin{equation}
    \begin{aligned}
      &L = L^{mse}_{full} + \lambda_1 L^{lpips}_{full} + \lambda_2 L_V,\\
      &L_V = L^{mse}_V + \lambda_{V_1} L^{lpips}_V + \lambda_{V_2} L^{dpt}_V,
      \label{eq:loss}
    \end{aligned}
\end{equation}
\end{small}where $\lambda_1$ and $\lambda_2$ are weights for LPIPS loss of $\mathcal{G}$ and composite loss of $\mathcal{G}_V$, respectively. $\lambda_{V_1}$ and $\lambda_{V_2}$ are weights for LPIPS and depth losses of $\mathcal{G}_V$, respectively.
%
\section{Experiments}
\label{sec:experiment}
\subsection{Experimental Setup}
\label{sec:experiment:setup}
We conduct experiments for both ego-centric and scene-centric sparse-view reconstruction tasks.
The ego-centric task is performed on nuScenes dataset \cite{nusc2020}, where the large motions and dense traffics in driving scenes pose more challenges for reconstruction.
\begin{table}[ht]
\centering
\resizebox{0.98\columnwidth}{!}{
\begin{tabular}{clccccc}
\hline
\toprule
\multicolumn{1}{l}{Dataset} & Method & PSNR$\uparrow$ & SSIM$\uparrow$ & LPIPS$\downarrow$ & PCC$\uparrow$ \\ \hline
\multicolumn{1}{l}{\multirow{3}{*}{nusc \cite{nusc2020}}} & pixelSplat \cite{pixelsplat2024} & 21.51 & 0.616 & 0.372 & 0.001 \\
\multicolumn{1}{l}{} & MVSplat \cite{mvsplat2024} & \underline{21.61} & \underline{0.658} & \underline{0.295} & \underline{0.181} \\
\multicolumn{1}{l}{} & Ours & \textbf{24.27} & \textbf{0.736} & \textbf{0.237} & \textbf{0.800} \\ \hline
\multirow{5}{*}{re10k \cite{re10k2018}} & AttnRend \cite{attnrend2023} & 24.78 & 0.820 & 0.213 & N/A \\
\multirow{5}{*}{} & MuRF \cite{murf2024} & 26.10 & 0.858 & 0.143 & 0.344 \\
\multirow{5}{*}{} & pixelSplat \cite{pixelsplat2024} & 25.89 & 0.858 & 0.142 & 0.285 \\
\multirow{5}{*}{} & MVSplat \cite{mvsplat2024} & \textbf{26.39} & \textbf{0.869} & \textbf{0.128} & \underline{0.363} \\
\multirow{5}{*}{} & Ours & \underline{26.19} & \underline{0.865} & \underline{0.131} & \textbf{0.368} \\
\bottomrule
\end{tabular}}
\vspace{-5pt}
\caption{Quantitative results on nuScenes \cite{nusc2020} and RealEstate10K \cite{re10k2018}. We \textbf{bold 1st-place} results and \underline{underline 2nd-place} results. PCC is not available (N/A) for light field-based method AttnRend which has no interpretable 3D structure for depth rendering.}
\label{tab:main_res}
\vspace{-14pt}
\end{table}
The scene-centric task is performed on RealEstate10K dataset \cite{re10k2018} following protocols presented in previous works \cite{pixelsplat2024,mvsplat2024}.

\noindent\textbf{Ego-Centric Task.}
The nuScenes dataset comprises 700 scenes for training and 150 scenes for validation, with each containing a video of approximately 20 seconds captured at 12 Hz.
We partition each scene into equally spaced bins along the vehicle trajectories, with a 3.2m interval between the first and the last captured frames. The central frame of each bin, featuring 6 surround-view images, serves as input views, while the first and the last frames, comprising 12 images, constitute the target novel views.
Thus, we obtain 135,941 bins used for training and 30,080 bins for validation in total. We adopt the image resolution of 224$\times$400 in our experiments for compatibility with the 2D diffusion model \cite{md2023}.
For evaluation, we compare our method against pixelSplat \cite{pixelsplat2024} and MVSplat \cite{mvsplat2024}, both are state-of-the-art methods for feed-forward sparse-view reconstruction.

\noindent\textbf{Scene-Centric Task.}
To further evaluate our method against prior works, we also conduct experiments on RealEstate10K, a large-scale scene-centric dataset containing both indoor and outdoor scenes.
Following the protocols adopted by previous works \cite{pixelsplat2024,mvsplat2024}, we use 67,477 scenes for training and 7,289 scenes for testing.
\begin{figure}[ht]
\centering
\vspace{-15pt}
\includegraphics[width=0.97\columnwidth]{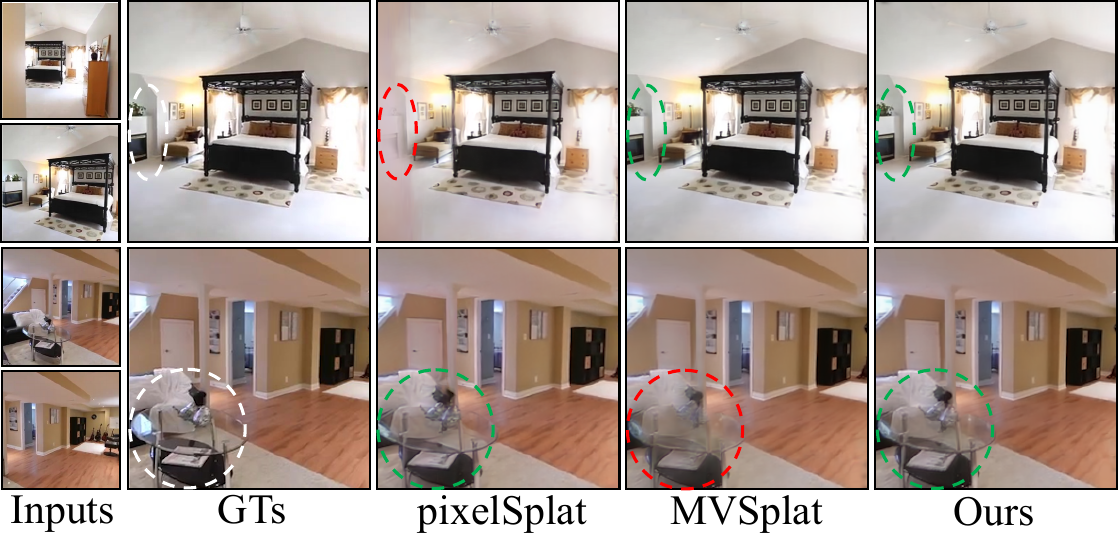}
\vspace{-5pt}
\caption{Comparisons on RealEstate10K \cite{re10k2018}. The \emph{red dashed circles} denote undesirable artifacts, while the \emph{green ones} denote plausibly-rendered areas.}
\label{Fig.fig4}
\vspace{-14pt}
\end{figure}
For evaluation, we conduct comprehensive comparisons with previous methods including 3DGS-based pixelSplat \cite{pixelsplat2024} and MVSplat \cite{mvsplat2024}, light field-based AttnRend \cite{attnrend2023}, and NeRF-based MuRF \cite{murf2024}. The results of these methods are adopted from their papers directly.

\noindent\textbf{Metrics.}
To measure visual quality, we use three metrics: peak signal-to-noise-ratio (PSNR), structural similarity (SSIM) \cite{ssim2004}, and perceptual distance (LPIPS) \cite{lpips2018}. For PSNR and SSIM, larger values are preferable, whereas the opposite for LPIPS.
To further assess geometric quality of 3D scenes, we compare rendered depth maps of novel views with those predicted by \cite{da2024}, which has been demonstrated to produce highly accurate and robust depth predictions in real-world scenarios.
Since \cite{da2024} can only obtain relative depths without scales, we use Pearson Correlation Coefficient (PCC) \cite{pcc2009} as the scale-invariant metric, which quantifies the statistical relationship between any two variables. The PCC ranges from -1 to 1, where -1 and 1 indicate perfect negative and positive relationships, respectively.

\noindent\textbf{Implementation Details.}
By default, the volume size $H{\times}W{\times}Z$ is set to 192$\times$192$\times$16, corresponding to the real world range of [-50m, -50m, -3m, 50m, 50m, 12m] around the vehicle.
The Triplane Transformer consists of three layers, with the first two incorporating both cross-image and cross-plane deformable attentions, while the last layer featuring only cross-plane deformable attention. The Volume Decoder adopts three linear layers to decode Gaussian parameters for voxel features.
For nuScenes dataset \cite{nusc2020}, our model is trained on two A100 GPUs for 100,000 iterations with the batch size of 4.
For RealEstate10K dataset \cite{re10k2018}, our model is trained on a single A100 GPU for 300,000 iterations with the batch size of 8.
The AdamW \cite{adam2014} optimizer is adopted with the learning rate of $1\times10^{-4}$ following cosine learning rate decay strategy.
More details can be found in our supplementary material.
\subsection{Main Results}
\label{sec:experiment:main}
\noindent\textbf{Ego-Centric Reconstruction.}
For evaluation, we make comparisons with state-of-the-art sparse-view reconstruction methods pixelSplat \cite{pixelsplat2024} and MVSplat \cite{mvsplat2024} re-implemented following their official code. They both adopt pixel-based Gaussian as the representation.
The quantitative results are shown in Table \ref{tab:main_res}. We can see that our method significantly surpasses others in terms of all metrics, especially for PCC that measures the geometric quality.
\begin{table}[h!t]
\centering
\vspace{-15pt}
\resizebox{1.0\columnwidth}{!}{
\begin{tabular}{lcccc}
\hline
\toprule
\multicolumn{1}{l}{Method} & PSNR$\uparrow$ & SSIM$\uparrow$ & LPIPS$\downarrow$ & PCC$\uparrow$ \\ \hline
\multicolumn{1}{l}{Volume-based} & {22.21} & {0.640} & {0.357} & {0.701} \\
\multicolumn{1}{l}{Pixel-based w/o depth init.} & {22.92} & {0.692} & {0.287} & {0.572} \\
\multicolumn{1}{l}{Pixel-based} & {22.89} & {0.698} & {0.290} & {0.780} \\ \hline
Full w/o train decomp. & {23.75} & {0.717} & {0.258} & {0.795} \\
Full w/o feat fuse. & {23.35} & {0.708} & {0.262} & {0.786} \\
Full & \textbf{24.27} & \textbf{0.736} & \textbf{0.237} & \textbf{0.800} \\
\bottomrule
\end{tabular}}
\vspace{-10pt}
\caption{Ablations on nuScenes \cite{nusc2020}. The 2nd to the 4th rows show results of models with only singular representations (volume or pixel-based Gaussian). The 5th to 7th rows show results of models with full Omni-Gaussian representation. Besides, ``depth init.", ``train decomp." and ``feat fuse." denote components of Depth Initialization for pixel-based Gaussian, Depth-Guided Training Decomposition and Projection-based Feature Fusion, respectively.}
\label{tab:abl_res}
\vspace{-10pt}
\end{table}
The main reason is that large cross-view overlap is unavailable for ego-centric reconstruction. Other methods cannot predict accurate depths using pixel-level 3D priors (e.g., epipolar lines \cite{pixelsplat2024} or cost volumes \cite{mvsplat2024}) that are dependent on cross-view correlation.
Besides, the drawbacks of pixel-based Gaussian also pose challenges for reconstruction.
Instead, we utilize volume-based Gaussian to lift 2D features to 3D space and predict Gaussians at the 3D level without relying on cross-view overlap.
Thanks to our dual-path deformable attentions, we can further mitigate the spatial limitations of pixel-based Gaussian during feature encoding.
The qualitative results are shown in Fig.\ref{Fig.fig3}.
We can see that both pixelSplat and MVSplat fail to render plausible depths, causing blurriness or inconsistency with ground truths in their results.
In contrast, our method can generate high-quality images and depths even if significant viewpoint changes exist between the input and the novel views.

\noindent\textbf{Scene-Centric Reconstruction.}
For evaluation, we compare our approach with more baseline methods on a widely-used scene-centric dataset (RealEstate10K \cite{re10k2018}).
As evident by the quantitative results in Table \ref{tab:main_res}, our method achieves comparable visual quality (measured by PSNR, SSIM and LPIPS) to state-of-the-art methods, and outperforms all prior works in terms of geometric quality (measured by PCC).
We also conduct qualitative comparisons in Fig.\ref{Fig.fig4}, where we can obtain novel views with better details than those produced by others, especially for cases of large motions.
Both of the quantitative and qualitative results show that our method not only exhibits superior performance in ego-centric reconstruction but also possesses competence in scene-centric reconstruction.

\noindent\textbf{Multi-Modal Generation.}
Our Omni-Scene can not only serve as a standalone reconstruction model, but also be seamlessly integrated with a 2D diffusion model \cite{md2023} to achieve feed-forward text-to-3D or layout-to-3D scene generation.
Specifically, given multi-modal conditions of textual descriptions or 3D layouts (e.g., 3D boxes, BEV map), we utilize \cite{md2023} to produce six single-frame surrounding images. Then, we can feed the images into our model to generate the corresponding explorable 3D scene with explicit 3D Gaussians.
\begin{figure*}[!ht]
\centering
\vspace{-17pt}
\includegraphics[width=0.9\textwidth]{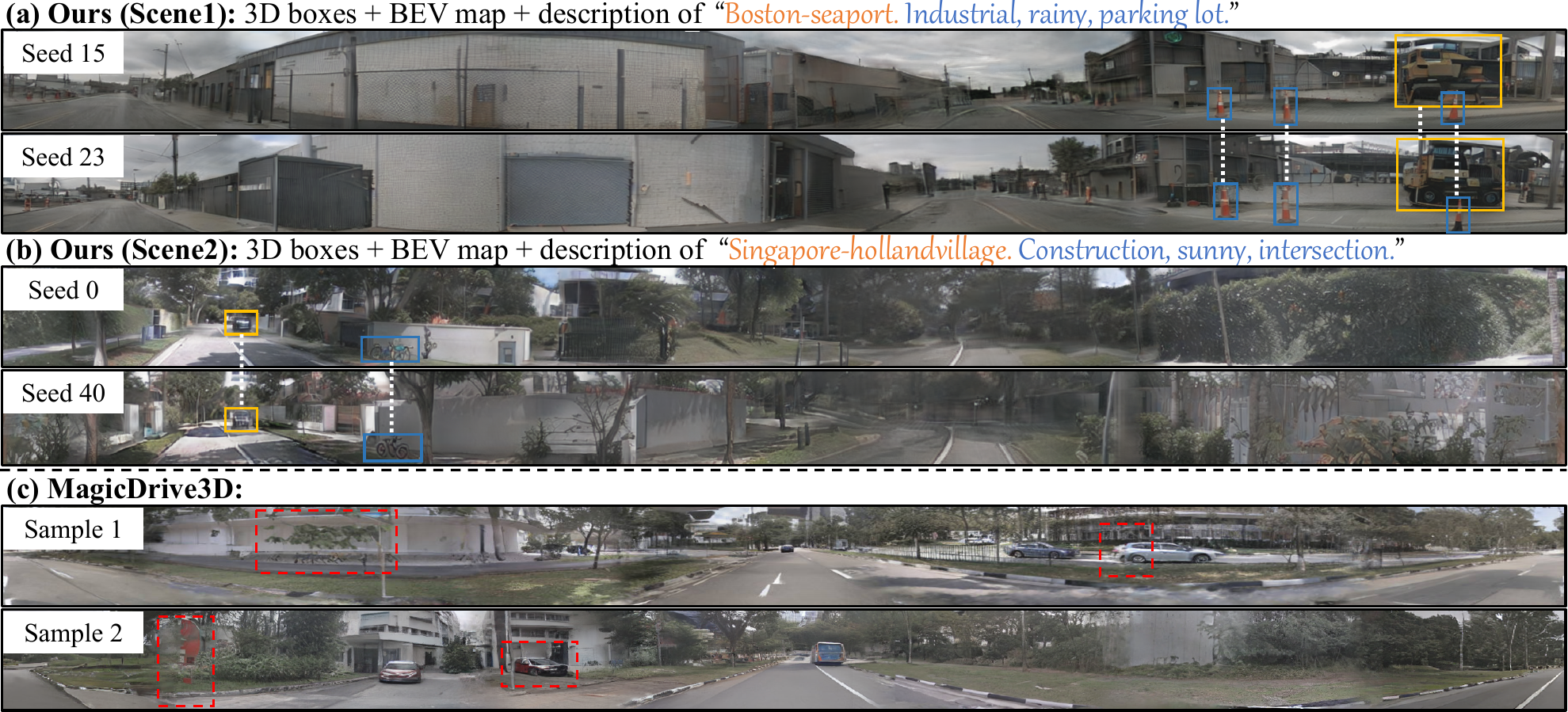}
\vspace{-5pt}
\caption{Multi-modal 3D scene generation. We accept multi-modal conditions (i.e., 3D boxes, BEV map, textual descriptions) as inputs, and generate the corresponding 3D driving scenes in a feed-forward manner. For better visualization, we render 360-degree rotation videos for the generated 3D scenes, and stitch frames into panoramic images as shown in (a) and (b). We can see that the styles of the generated scenes closely match the textual conditions.
Besides, when the appearances vary with random seeds, the spatial consistency with conditional 3D boxes (denoted by colored rectangles in (a) and (b)) is well preserved. Compared to per-scene optimization-based method MagicDrive3D \cite{md3d2024} that leads to artifacts highlighted by red dashed lines in (c), we achieve higher quality with better visual details. \emph{Please consult our supplementary material for comparisons in video format}, where we can better observe the differences in visual quality.}
\label{Fig.fig6}
\vspace{-13pt}
\end{figure*}
The most relevant work to us is MagicDrive3D \cite{md3d2024}, which first uses video diffusion model to generate multi-view video with approximately 100 images, and then reconstruct the scene based on deformable Gaussians \cite{d3dgs2024}. Such scene-by-scene reconstruction is inefficient and demands high spatial and temporal consistency from the generated videos, often failing and introducing noise or jitter artifacts in the synthetic 3D scenes.
As shown in Fig.\ref{Fig.fig6}(a)-(b), our generated results exhibit high fidelity and good diversity, while also ensure consistency with both the textual and layout conditions.
Compared to results from MagicDrive3D \cite{md3d2024} as shown in Fig.\ref{Fig.fig6}(c), we achieve better quality in a much more efficient feed-forward manner.
This demonstrates the potential of Omni-Scene for generation, pioneering a new approach to multi-modal generation of 3D driving scenes in a feed-forward manner.

\begin{figure}[hb]
\vspace{-15pt}
\centering
\includegraphics[width=0.98\columnwidth]{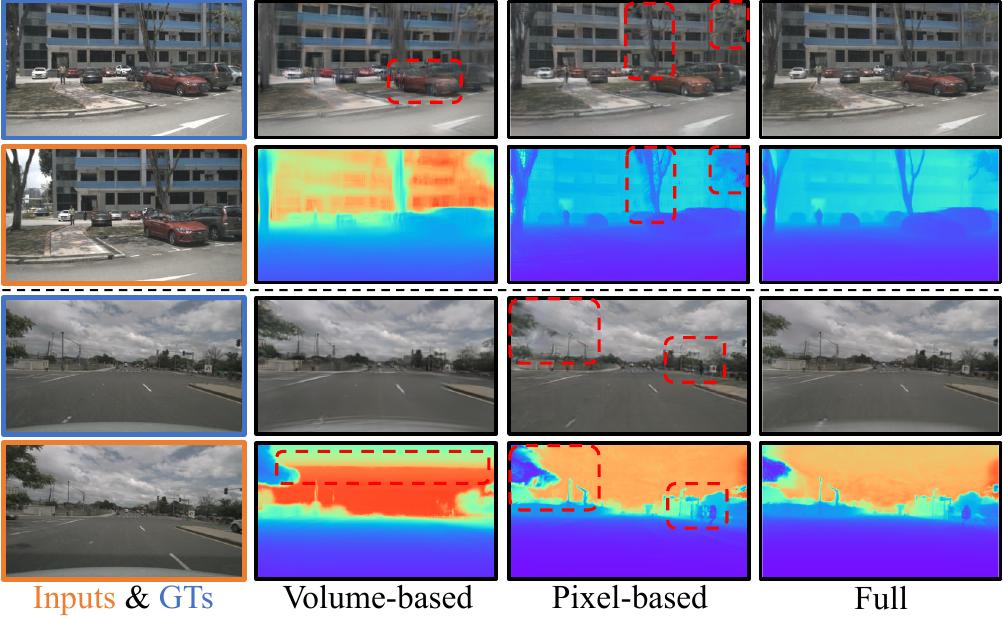}
\vspace{-5pt}
\caption{Ablations on Omni-Gaussian representation. Images of input views (Inputs) and ground-truth novel views (GTs) are outlined by orange and blue rectangles, respectively.}
\label{Fig.fig5}
\end{figure}
\noindent\textbf{Additional Results.}
\emph{Please refer to our supplementary material} for additional results including \textbf{scene-exploring videos}, runtime analysis, more comparisons, etc.
\subsection{Ablation Study}
\label{sec:experiment:abl}
\noindent\textbf{Effectiveness of Omni-Gaussian Representation.}
The core of our method lies in the Omni-Gaussian representation.
Therefore, we train two variant models with only volume or pixel-based Gaussian representation for comparisons.
By comparing our ``Full" method with ``Volume-based" and ``Pixel-based" in Table \ref{tab:abl_res}, we can see that the full method with Omni-Gaussian representation surpasses the two singular representation variants by a large margin.
Without volume-based Gaussian to address object occlusions and frustum truncations, we can observe a performance drop of 1.38dB PSNR, 0.038 of SSIM, and 0.02 of PCC.
Removing pixel-based Gaussian which refines details and reconstructs distant elements, PSNR, SSIM and PCC will be decreased by 2.06dB, 0.096 and 0.099, respectively.
Such deterioration is more evident in Fig.\ref{Fig.fig5}.
With only volume-based Gaussian, we observe clear depth boundaries (i.e., last row, 2nd column of Fig.\ref{Fig.fig5}) and lack of visual details (i.e., 1st row, 2nd column of Fig.\ref{Fig.fig5}), corresponding to Case 3 and 4 in Fig.\ref{Fig.fig1}(a). With only pixel-based Gaussian, we observe noise artifacts in areas occluded or truncated in 2D (3rd column of Fig.\ref{Fig.fig5}), corresponding to Case 1 and 2 in Fig.\ref{Fig.fig1}(a). With the collaboration of the two representations, our full method can eliminate these artifacts and achieve best consistency with GTs (4th column of Fig.\ref{Fig.fig5}).
Moreover, we show examples in Fig.\ref{Fig.fig1}(b)-(e) to further illustrate benefits from such collaboration.

\noindent\textbf{Effectiveness of Volume-Pixel Collaboration.}
As shown in Table \ref{tab:abl_res}, our ``Full" method outperforms the variants ``Full w/o feat fuse." and ``Full w/o train decomp." in terms of all metrics.
This indicates that the collaboration between volume and pixel-based representations is important in both the feature encoding stage and the training stage.
The Projection-Based Feature Fusion strategy enables our Volume Builder to be aware of which areas are already covered by pixel-based Gaussian, allowing it to better complement the uncovered areas.
The Depth-Guided Training Decomposition mechanism allows our Volume Builder to focus on reconstruction within the volume range, while also ensuring the spatial alignment between the volume and the pixel-based Gaussians, which avoids scale ambiguity.
Visual results can be found in our supplementary material.

\noindent\textbf{Effectiveness of Depth Initialization.}
We also train a pixel-based variant model without initializing per-pixel depths using \cite{metric3d2024} to investigate the impact of depth initialization on performance.
As shown in Table \ref{tab:abl_res}, although the variant ``Pixel-based w/o depth init." can achieve comparable visual quality to the full-version ``Pixel-based", it leads to a 0.208 drop of PCC.
The main reason for this gap is that the depth initialization can ease the prediction of complex 3D geometries under the ego-centric setting.
Visual results can be found in our supplementary material.

\section{Conclusion}
We have introduced Omni-Scene, a method with Omni-Gaussian representation that can reach the best of both pixel and volume-based Gaussian representations for ego-centric sparse-view scene reconstruction.
Employing designs that encourage Volume-Pixel collaboration, we achieve high-fidelity scene reconstruction from only single-frame surrounding observations.
Extensive experiments demonstrate our superiority in ego-centric reconstruction compared to previous methods.
Furthermore, we integrate a 2D diffusion model into our framework, which enables multi-modal 3D scene generation with versatile applications.
{
    \small
    \bibliographystyle{unsrt}
    \bibliography{main}
}

\maketitlesupplementary
\setcounter{page}{1}
In this document, we first provide implementation details including data preprocessing of nuScenes \cite{nusc2020} (Sec.\ref{sec:supl:data}), network architecture and hyperparameters (Sec.\ref{sec:supl:detail}).
We follow with additional experiment discussions including introduction to our supplementary videos (Sec.\ref{sec:supl:video}), quantitative comparisons with more baseline methods (Sec.\ref{sec:supl:comp}), runtime analysis (Sec.\ref{sec:supl:runtime}), more ablations (Sec.\ref{sec:supl:abl}), further discussions on generalizability to larger bins (Sec. \ref{sec:supl:bin}) and effectiveness of our Volume-Pixel Collaboration (Sec. \ref{sec:supl:col}), more qualitative results on scene-centric reconstruction (Sec.\ref{sec:supl:re10k}).
\emph{\textbf{We strongly recommend to view the accompanying video (``video.mp4")}}, which contains 360-degree exploring videos of both reconstructed and synthetic scenes, as well as comparisons with other methods.

\section{Additional Implementation Details}
\subsection{Data Preprocessing}
\label{sec:supl:data}
\begin{figure}[hb]
\centering
\includegraphics[width=\columnwidth]{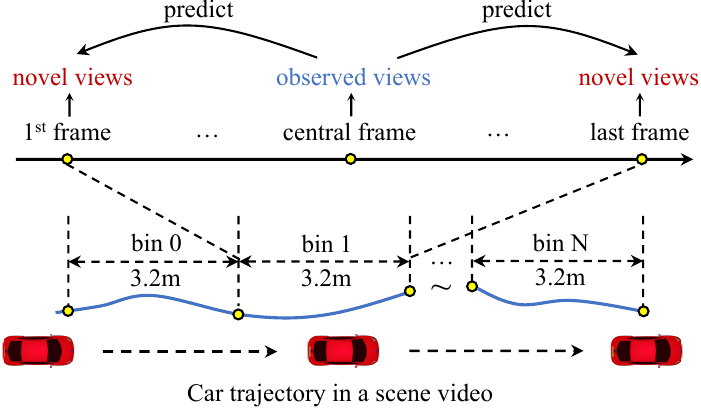}
\caption{Data preprocessing of nuScenes \cite{nusc2020}.}
\label{Fig.supl_data}
\end{figure}

As described in Sec.\ref{sec:experiment:setup} of our main manuscript, we partition each scene of nuScenes dataset \cite{nusc2020} into equally spaced bins, with each bin serving as one data sample.
For nuScenes dataset, each video is captured in a single scene along with the car trajectory.
The length of the trajectory ranges drastically from several meters to hundreds of meters.
If we segment the trajectory into bins according to frame indexes, the spatial ranges of bins would exhibit significant variation, which leads to non-IID data distribution for training and evaluation.
To circumvent this issue, we segment the bins based on the distance traveled by the car as detailed in Fig.\ref{Fig.supl_data}.
Specifically, for videos with a trajectory length exceeding 3.2 meters, we uniformly segment them into $N$ bins, each 3.2 meters in length.
For each bin, we use the central frame with 6 surrounding images
to derive the observed input views, and the first and last frames with 12 surrounding images as the novel views.
For videos with a trajectory length less than 3.2 meters, we directly use the first and last frames of the video as the novel views.

\subsection{Network Architecture and Hyperparameters}
\label{sec:supl:detail}
In Table \ref{tab:supl:detail}(a), the order from top to bottom are the parameters of Triplane Transformer (i.e., number of transformer layers, embedding dimensions, number of 2D and 3D reference points used in our cross-image and cross-plane deformable attentions, and number of attention heads), Voxel Decoder (i.e., number of Gaussians decoded for each voxel, number of linear layers used for decoding Gaussian parameters), Multi-View U-Net (i.e., feature dimensions and patch sizes of patchified cross attentions \cite{pixart2024} used in U-Net downsample and upsample blocks), Pixel Decoder (i.e., number of convolution layers used for decoding Gaussian parameters), respectively.
In Table \ref{tab:supl:detail}(b), we specify loss weights for Eq.(4) in our main manuscript, which is followed by parameters used in our training phase.

\begin{table}[h!t]
\centering
\resizebox{\columnwidth}{!}{
\begin{tabular}{lll}
\toprule
\multicolumn{3}{c}{(a) Network Architecture} \\ \hline
\specialrule{0em}{1pt}{0pt}
\multirow{2}{*}{2D Image Encoder} & backbone & R50-DINO \cite{dino2021} \\
\multirow{2}{*}{} & neck & FPN (P2 only) \cite{fpn2017} \\ \hline
\specialrule{0em}{1pt}{0pt}
\multirow{5}{*}{Triplane Transformer} & \# layers & 3 \\
\multirow{5}{*}{} & \# embed dims & 128 \\
\multirow{5}{*}{} & \# 2D ref points & 8, 16, 16 \\
\multirow{5}{*}{} & \# 3D ref points & 16, 16, 16 \\
\multirow{5}{*}{} & \# attn heads & 8 \\ \hline
\multirow{2}{*}{Voxel Decoder} & \# Gaussians per voxel & 3 \\
\multirow{2}{*}{} & \# linear layers & 3 \\ \hline
\multirow{4}{*}{Multi-View U-Net} & \# downsample feats & 128, 256, 512, 512 \\
\multirow{4}{*}{} & \# upsample feats & 512, 512, 256, 128 \\
\multirow{4}{*}{} & \# downsample patches & 8, 8, 4, 2 \\
\multirow{4}{*}{} & \# upsample patches & 2, 4, 8, 8 \\ \hline
\specialrule{0em}{1pt}{0pt}
Pixel Decoder & \# conv layers & 3 \\
\toprule
\specialrule{0em}{3pt}{0pt}
\multicolumn{3}{c}{(b) Hyperparameters} \\
\specialrule{0em}{1pt}{0pt} \hline
\specialrule{0em}{1pt}{1pt}
Loss Weights & \# $\lambda_1,\lambda_2,\lambda_{V_1},\lambda_{V_2}$ & 0.05, 1.0, 0.05, 0.01 \\ \hline
\specialrule{0em}{1pt}{0pt}
\multirow{8}{*}{Training Details} & learning rate scheduler & Cosine \\
\multirow{8}{*}{} & \# iterations & 100,000 \\
\multirow{8}{*}{} & \# learning rate & 1e-4 \\
\multirow{8}{*}{} & optimizer & Adam \cite{adam2014} \\
\multirow{8}{*}{} & \# beta1, beta2 & 0.9, 0.999 \\
\multirow{8}{*}{} & \# weight decay & 0.01 \\
\multirow{8}{*}{} & \# warm-up & 1000 \\
\multirow{8}{*}{} & \# gradient clip & 1.0 \\
\bottomrule
\end{tabular}}
\caption{Details of network architecture and hyperparameters. In the table, ``\#'' denotes numerical parameters. We present parameters that specify our network architecture, and parameters used in our loss functions and training phase, in (a) and (b), respectively.}
\label{tab:supl:detail}
\vspace{-10pt}
\end{table}
\begin{table*}[h!t]
\vspace{-20pt}
\centering
\resizebox{0.6\textwidth}{!}{
\begin{tabular}{lcccccc}
\hline
\toprule
Method & Time(s) & Param(M) & PSNR$\uparrow$ & SSIM$\uparrow$ & LPIPS$\downarrow$ & PCC$\uparrow$ \\ \hline
AttnRend \cite{attnrend2023} & {9.98} & {125.1} & {20.96} & {0.533} & {0.467} & {N/A} \\
MuRF \cite{murf2024} & {0.672} & {\textbf{5.3}} & {20.34} & {0.504} & {0.433} & {-0.332} \\
pixelSplat \cite{pixelsplat2024} & {0.508} & {125.4} & {21.51} & {0.616} & {0.372} & {0.001} \\
MVSplat \cite{mvsplat2024} & {\underline{0.174}} & {\underline{12.0}} & {\underline{21.61}} & {\underline{0.658}} & {\underline{0.295}} & {\underline{0.181}} \\
Ours & {\textbf{0.088}} & {81.7} & {\textbf{24.27}} & {\textbf{0.736}} & {\textbf{0.237}} & {\textbf{0.800}} \\
\bottomrule
\end{tabular}}
\vspace{-5pt}
\caption{Additional quantitative results on ego-centric reconstruction task performed on nuScenes \cite{nusc2020}. We \textbf{bold 1st-place} results and \underline{underline 2nd-place} results. PCC is not available (N/A) for AttnRend which has no interpretable 3D structure for depth rendering.}
\label{tab:supl:main_res}
\vspace{-10pt}
\end{table*}

\section{Additional Experiments}
\subsection{Video Results}
To better demonstrate the quality of 3D reconstruction, we provide \textbf{exploring video demos in ``video.mp4"} along with our supplementary material.
Specifically, given six surrounding images of a scene, we conduct inference and obtain 3D Gaussians for reconstructing the scene.
Then, we utilize these Gaussians to render a 360-degree rotation video at 30fps with the camera FOV set to 70 degree following \cite{nusc2020}.
In the video, each frame that falls between the input viewpoints can be considered as a novel view unseen in the inputs.
To further demonstrate the model's performance in the presence of object occlusions and frustum truncations, we move the camera forward and backward by 3 meters in the front and rear view perspectives, respectively, ensuring that there are contents invisible from the input views.
It's also noted that the camera's movement range has reached 6 meters, exceeding the 3.2-meter range of bin samples seen during training, thereby showcasing the model's capability to reconstruct scenes at greater distances.

\label{sec:supl:video}
\noindent\textbf{Comparisons with other methods.}
We first present comparisons with state-of-the-art methods pixelSplat \cite{pixelsplat2024} and MVSplat \cite{mvsplat2024} from \emph{00:00 to 01:40 in ``video.mp4"}. Our approach significantly outperforms other methods in both visual and geometric quality. Notably, due to the minimal cross-view overlap among input views, pixelSplat and MVSplat fail to predict accurate depths based on pixel-level 3D priors (e.g., epipolar lines, cost volumes), which results in artifacts in the rendered videos especially when the camera is substantially moved forward and backward.

\noindent\textbf{Exploring videos of reconstructed scenes.}
Then, we present more examples to illustrate our functionality on scene reconstruction.
Examples with normal conditions are shown from \emph{01:41 to 02:49 in ``video.mp4"}.
Examples with extreme conditions (e.g., low-light, bad weather) are shown from \emph{02:50 to 03:26 in ``video.mp4"}. We can see that our method achieves high-quality reconstruction and maintains robustness in both normal and hard cases.

\noindent\textbf{Exploring videos of generated scenes.}
We also present examples to illustrate our functionality on scene generation from \emph{03:27 to 05:07 in ``video.mp4"}.
The left side of the video shows the our generated results given different random seeds.
The right side of the video shows examples of MagicDrive3D \cite{md3d2024}, which are directly adopted from their official website\footnote{https://gaoruiyuan.com/magicdrive3d/}.
We can see that our method achieves better visual details than per-scene optimization-based MagicDrive3D in a much more efficient feed-forward manner.

\subsection{Comparisons with More Baselines}
\label{sec:supl:comp}
We also make comparisons with more baseline methods (i.e., MuRF \cite{murf2024} and AttnRend \cite{attnrend2023}) for ego-centric sparse-view reconstruction task.
Specifically, MuRF and AttnRend are feed-forward reconstruction methods based on NeRF \cite{nerf2021} and light field \cite{lf2022}, respectively.
They are both leading and representative methods within their respective lines of works, which constitute the mainstream feed-forward methods together with 3DGS-based approaches such as pixelSplat \cite{pixelsplat2024} and MVSplat \cite{mvsplat2024}.
As shown in Table \ref{tab:supl:main_res}, our method surpasses MuRF and AttnRend significantly in terms of all metrics.
We can also observe that methods with explicit Gaussians as representations (i.e., ours, pixelSplat, MVSplat) outperform methods with implicit NeRF or light field as representations (i.e., AttnRend, MuRF), showing the effectiveness of explicit 3D representation.

\subsection{Runtime Analysis}
\label{sec:supl:runtime}
As shown in Table \ref{tab:supl:main_res}, we conduct runtime analysis on the ego-centric reconstruction task to demonstrate the efficiency of our method.
It's noted that the inference speed is reported based on the time cost of six-view reconstruction averaged by 2,048 times.
From the table, we can see that our method achieves the shortest inference time (i.e., ``Time" in Table \ref{tab:supl:main_res}), which is nearly 2$\times$ faster than that of the 2nd place method MVSplat \cite{mvsplat2024}.
We attribute this advantage to our triplane-based volume feature encoding in Triplane Transformer, and efficient patchified cross-attention module in Multi-View U-Net.
Besides, our method is also light-weight with model size (i.e., ``Param'' in Table \ref{tab:supl:main_res}) comparable to other methods.
Furthermore, we observe that, thanks to the efficient rendering of 3DGS \cite{3dgs2023}, 3DGS-based methods (i.e., our method, pixelSplat \cite{pixelsplat2024}, MVSplat \cite{mvsplat2024}) show significant superiority in speed compared to methods based on implicit representations (i.e., MuRF \cite{murf2024}, AttnRend \cite{attnrend2023}).

\subsection{Additional Ablations}
\label{sec:supl:abl}
We present more ablation results to demonstrate the effectiveness of our components.

\noindent\textbf{Qualitative Ablations on Volume-Pixel Collaboration.}
\begin{figure*}[!ht]
\centering
\includegraphics[width=0.9\textwidth]{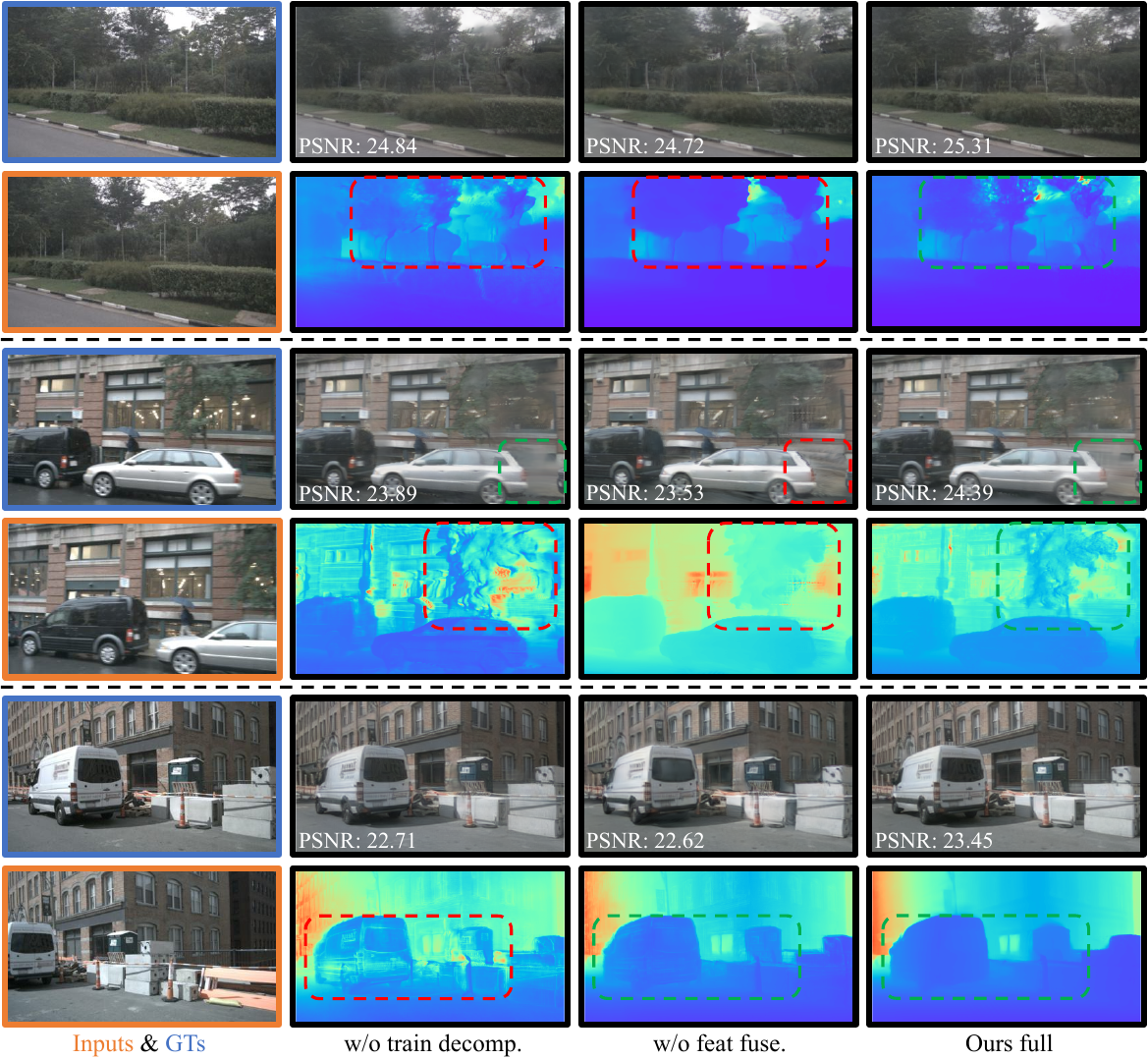}
\vspace{-3pt}
\caption{Qualitative ablations on Volume-Pixel Collaboration. Images of input views (Inputs) and ground-truth novel views (GTs) are outlined by orange and blue rectangles, respectively. The remaining are novel-view images and depths generated by our variant models and full model. From left to right, the order is the variant without Depth-Guided Training Decomposition, the variant without Projection-Based Feature Fusion, and our full method. The \emph{red dashed lines} highlight undesirable artifacts (e.g., noise, over-smooth), while the \emph{green ones} denote plausibly-rendered areas (e.g., better and sharper details). We also show PSNR values of the generated images for better comparison.}
\label{Fig.abl_col}
\vspace{-10pt}
\end{figure*}
In our main manuscript, we have quantitatively compared our full method with the two variants without the Volume-Pixel Collaboration designs (i.e., Projection-based Feature Fusion and Depth-Guided Training Decomposition).
Here, we show additional qualitative results in Figure \ref{Fig.abl_col} for visual comparisons.
It can be observed that our full method can generate images with higher quality and depths with sharper details, which demonstrate that our collaboration designs can effectively encourage the complementarity between pixel-based and volume-based Gaussian representations, and further improve the performance.

\begin{figure*}[!ht]
\centering
\vspace{-15pt}
\includegraphics[width=0.88\textwidth]{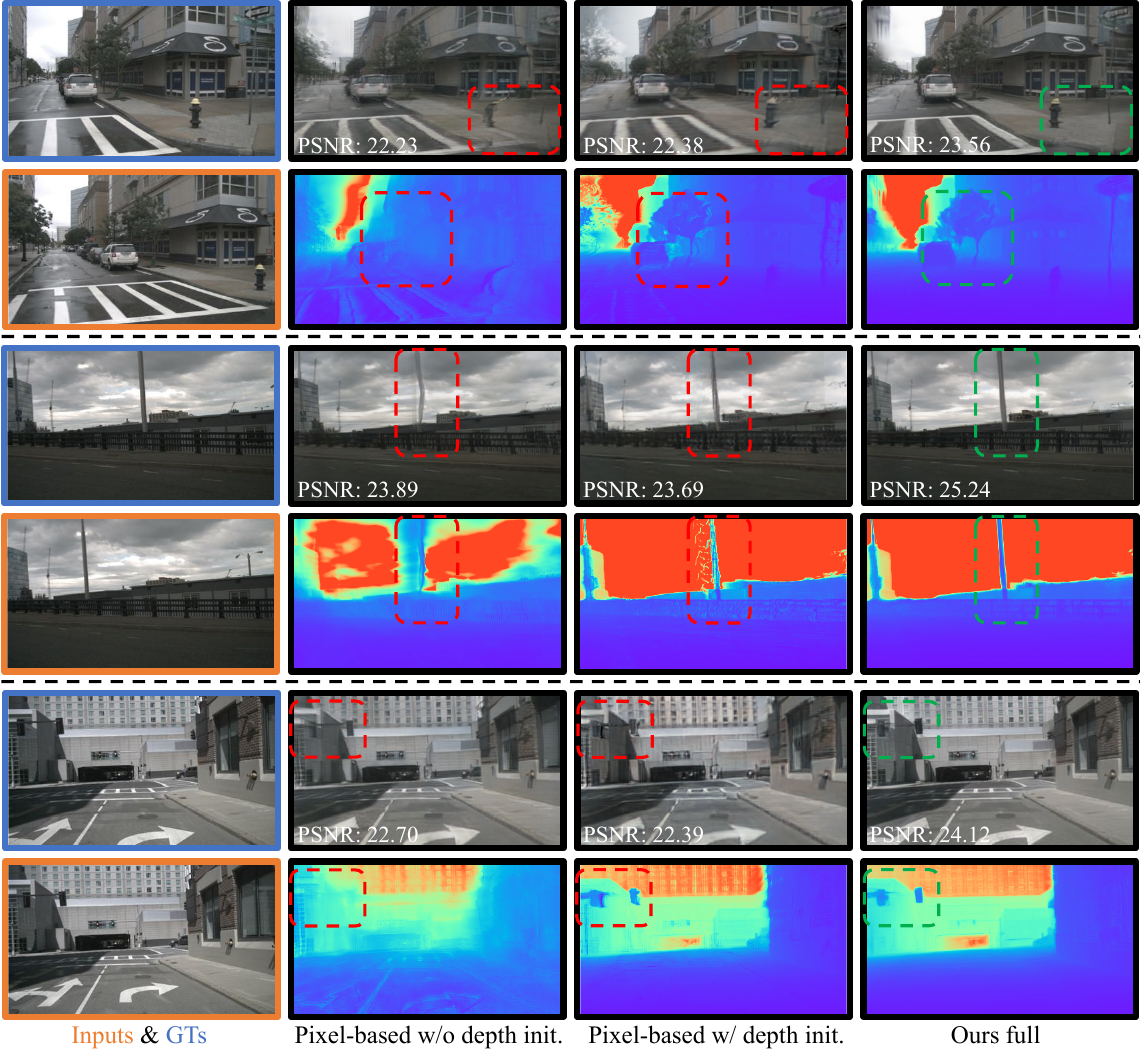}
\vspace{-5pt}
\caption{Qualitative ablations on Depth Initialization. The 1st column present images of input views (Inputs) and ground-truth novel views (GTs). The 2nd and the 3rd columns are results generated by two variant models with only pixel-based representation (i.e., Pixel Decorator), one with the depth initialization and one without. The last column denote results generated by our full method. The \emph{red dashed lines} highlight undesirable artifacts, while the \emph{green ones} denote plausibly-rendered areas. PSNR values are shown for better comparison.}
\label{Fig.abl_dpt}
\vspace{-10pt}
\end{figure*}

\noindent\textbf{Qualitative Ablations on Depth Initialization.}
In our main manuscript, we have quantitatively demonstrate the effectiveness of depth initialization for our pixel-based Gaussian representation.
Here, we show additional qualitative results in Figure \ref{Fig.abl_dpt} for visual comparisons.
From the figure, we can see that, although the depth initialization has no significant impact on visual quality, it is beneficial for improving geometric quality. The main reason is that the depth initialization can ease the learning of complex scene geometries for our Pixel Decorator that built upon pixel-based representation.
Besides, with the collaboration of volume-based representation, our full method significantly surpasses the two variants with only pixel-based representations both visually and geometrically, further demonstrating the advantage of our Omni-Gaussian representation.

\noindent\textbf{Ablations on Deformable Attentions.}
As described in Sec.\ref{sec:method:vol} of our main manuscript, we employ cross-image and cross-plane deformable attentions in our Volume Builder to enhance volumetric feature encoding.
Given camera parameters (i.e., intrinsics and extrinsics) that enable 3D-to-2D projection, our cross-image deformable attention module can lift 2D features to the 3D volume space, which enables the prediction of 3D Gaussians directly at the 3D level.
This differs from previous methods \cite{pixelsplat2024,mvsplat2024} that require cross-view overlap to estimate per-pixel depths and predict 3D Gaussians at the 2D level.
\begin{table}[!hb]
\centering
\resizebox{\columnwidth}{!}{
\vspace{-15pt}
\begin{tabular}{cccccc}
\hline
\toprule
cross-image & cross-plane & PSNR$\uparrow$ & SSIM$\uparrow$ & LPIPS$\downarrow$ & PCC$\uparrow$ \\ \hline
\ding{55} & \ding{51} & {14.29} & {0.428} & {0.578} & {0.539} \\
\ding{51} & \ding{55} & {21.29} & {0.595} & {0.412} & {0.686} \\
\ding{51} & \ding{51} & {\textbf{22.21}} & {\textbf{0.640}} & {\textbf{0.357}} & {\textbf{0.701}} \\
\bottomrule
\end{tabular}}
\vspace{-5pt}
\caption{Ablations on cross-image \& plane deformable attentions.}
\label{tab:supl:abl_attn}
\end{table}
To further address the issue that some elements in 3D might be occluded or truncated for any of the 2D input views, we utilize our cross-plane deformable attention to enhance each triplane query with cross-plane context, which means information absent in one plane can be complemented by those from other planes at the 3D level.
To validate the effectiveness of such dual-path design, we train three Volume Builder models, where one contains both of the cross-image and cross-plane attentions, while the other two contain only one of the attentions.
\begin{figure*}[!ht]
\centering
\vspace{-5pt}
\includegraphics[width=\textwidth]{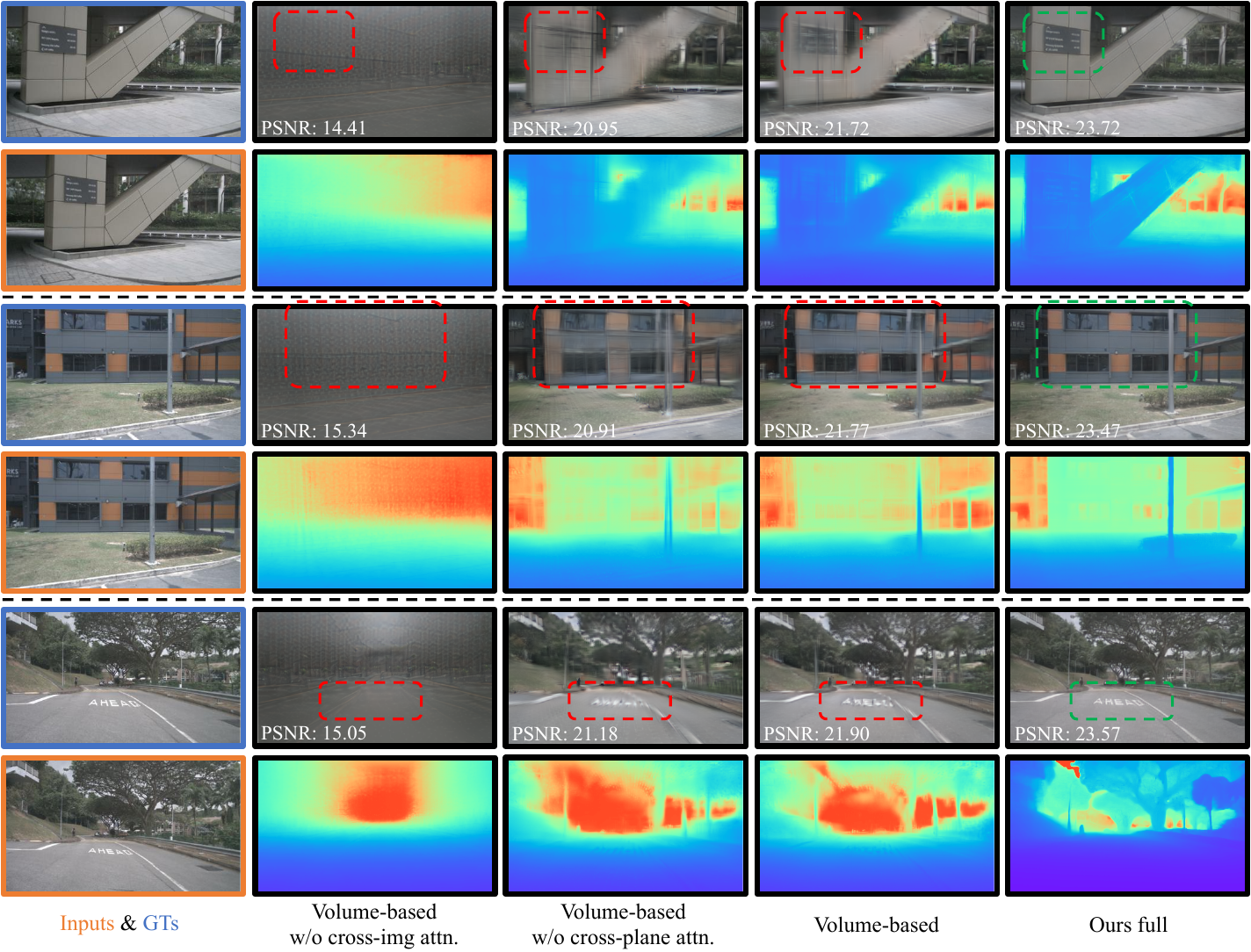}
\caption{Qualitative ablations on deformable attentions. The 1st column present images of input views (Inputs) and ground-truth novel views (GTs). The 2nd to 4th columns are results generated by three variant models with only volume-based representation (i.e., Volume Builder), one without the cross-image deformable attention (``cross-img attn."), one without the cross-plane deformable attention (``cross-plane attn."), and one with both of the attentions. The last column denote results generated by our full method. The \emph{red dashed lines} highlight undesirable artifacts, while the \emph{green ones} denote plausibly-rendered areas. PSNR values are shown for better comparison.}
\label{Fig.abl_attn}
\end{figure*}
As demonstrated in Table \ref{tab:supl:abl_attn} and Fig. \ref{Fig.abl_attn}, the model with both attentions significantly outperforms the other two variants, showing the importance of such dual-path feature encoding to our Volume Builder.
We further compare these volume-only variants with our full method in Fig. \ref{Fig.abl_attn}.
It's observed that results generated by our full method are with better details, showing the effectiveness of our Omni-Gaussian representation.

\subsection{Generalizability to Different Bin Sizes}
\label{sec:supl:bin}
Unless otherwise specified, our experiments are conducted with a bin size of 3.2m as stated in Sec.\ref{sec:supl:data}.
To validate whether our method can be generalized to synthesize novel views at farther or closer distances, we preprocess nuScenes \cite{nusc2020} into three variants with different bin sizes (i.e., 1.6m, 6.4m, 12.8m) from our original dataset.
Here we note that, the larger the bin size, the farther distance between the novel and the input views, which is more challenging for novel view synthesis.
Practically, for each dataset variant, we employ two approaches to test our model: (1) The model trained under a bin size of 3.2m is directly used for evaluation without additional fine-tuning. (2) The model is further fine-tuned with the new bin size for 50,000 steps before evaluation.
As can be seen from the 3rd, 5th and 7th rows of Table \ref{tab:supl:abl_bin}, despite the lack of supervision, our method exhibits minor degradation in performance for novel view synthesis at farther distances. For instance, we observe only 1.38 dB drop of PSNR, and 0.009 drop of PCC for ``bin size = 6.4m", which denotes distances 2$\times$ farther than those seen during training.
\begin{table*}[!ht]
\vspace{-17pt}
\centering
\resizebox{0.6\textwidth}{!}{
\vspace{-15pt}
\begin{tabular}{cccccc}
\hline
\toprule
bin size & fine-tuning & PSNR$\uparrow$ & SSIM$\uparrow$ & LPIPS$\downarrow$ & PCC$\uparrow$ \\ \hline
3.2m & -- & {24.27} & {0.736} & {0.237} & {0.800} \\ \hline
\specialrule{0em}{2pt}{0pt}
\multirow{2}{*}{1.6m} & \ding{55} & {25.12$^{+0.85}$} & {0.771$^{+0.035}$} & {0.208$^{-0.030}$} & {0.804$^{+0.004}$} \\
\multirow{2}{*}{} & \ding{51} & {25.37$^{+1.10}$} & {0.783$^{+0.047}$} & {0.201$^{-0.037}$} & {0.806$^{+0.006}$} \\ \hline
\specialrule{0em}{2pt}{0pt}
\multirow{2}{*}{6.4m} & \ding{55} & {22.89$^{-1.38}$} & {0.682$^{-0.054}$} & {0.287$^{+0.050}$} & {0.791$^{-0.009}$} \\
\multirow{2}{*}{} & \ding{51} & {24.15$^{-0.12}$} & {0.729$^{-0.007}$} & {0.239$^{+0.002}$} & {0.797$^{-0.003}$} \\ \hline
\specialrule{0em}{2pt}{0pt}
\multirow{2}{*}{12.8m} & \ding{55} & {21.57$^{-2.70}$} & {0.640$^{-0.096}$} & {0.346$^{+0.109}$} & {0.771$^{-0.029}$} \\
\multirow{2}{*}{} & \ding{51} & {23.55$^{-0.72}$} & {0.711$^{-0.025}$} & {0.265$^{+0.028}$} & {0.792$^{-0.008}$} \\
\specialrule{0em}{-1pt}{0pt}
\bottomrule
\end{tabular}}
\vspace{-8pt}
\caption{Results of our method when generalized to different bin sizes with or without additional fine-tuning.}
\label{tab:supl:abl_bin}
\end{table*}
\begin{figure*}[!ht]
\centering
\includegraphics[width=\textwidth]{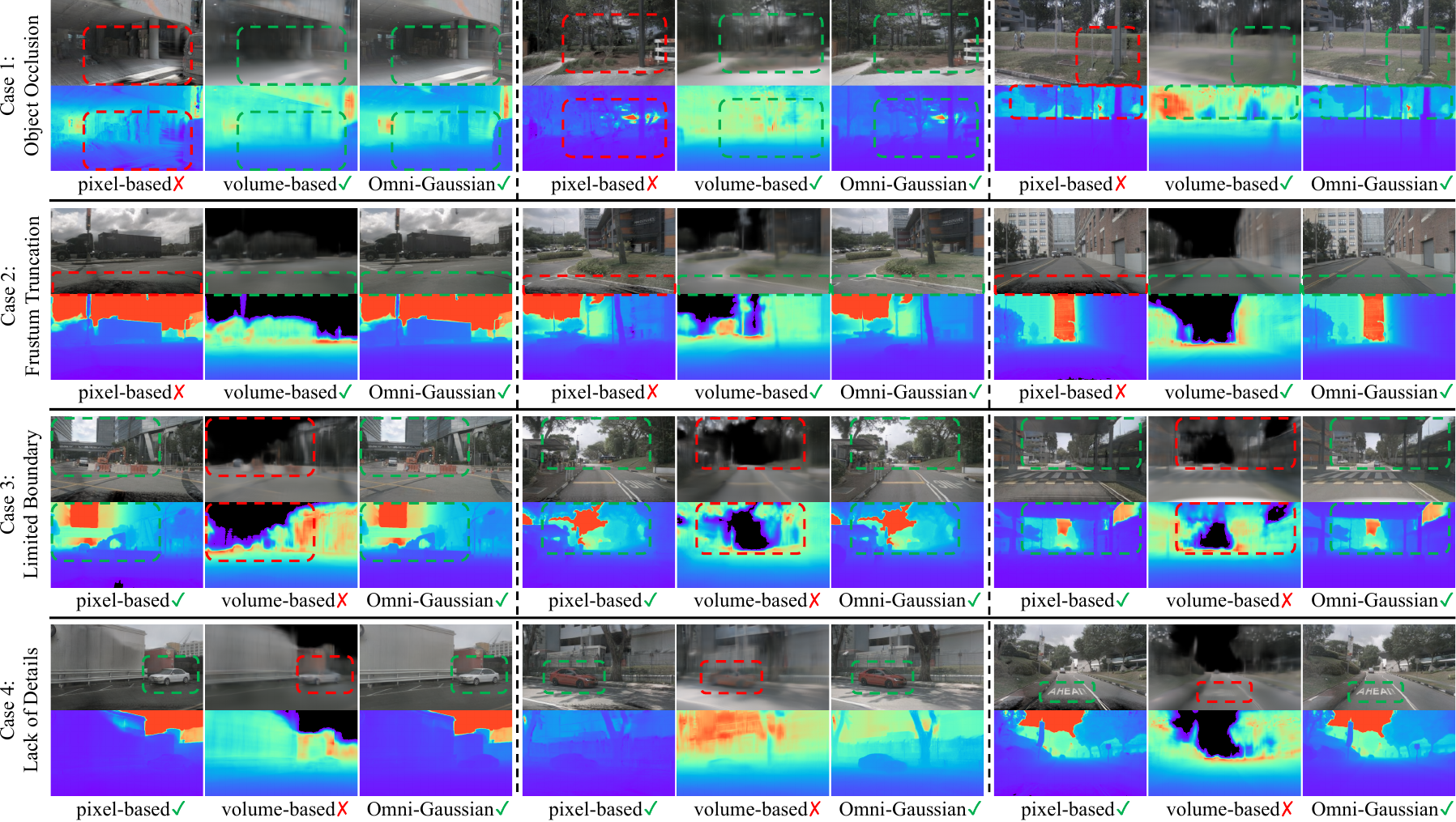}
\vspace{-5pt}
\caption{Additional examples of Volume-Pixel Collaboration. The \emph{red dashed lines} highlight artifacts caused by weaknesses of singular representations, while the \emph{green ones} outline how the artifacts are eliminated through Volume-Pixel Collaboration.}
\label{Fig.example}
\vspace{-10pt}
\end{figure*}
As can be seen from the 4th, 6th, and 8th rows of Table \ref{tab:supl:abl_bin}, by fine-tuning the model on data renewed with different bin sizes, we can further boost the performance and bring novel view synthesis at farther distances (i.e., ``bin size = 6.4m, 12.8m") very close to the results obtained under the original setting of ``bin size = 3.2m".

\subsection{Discussion on Volume-Pixel Collaboration}
\label{sec:supl:col}
In Fig.\ref{Fig.fig1} of our main manuscript, we have showcased pros and cons of the pixel-based and the volume-based Gaussian representations, and have provided the corresponding examples to illustrate \emph{how the two representations complement for each other in our unified model with the proposed Omni-Gaussian representation}.
Here, we present more examples in Fig.\ref{Fig.example} to demonstrate the effectiveness of their collaboration case by case:
\begin{itemize}
\item In ``Case 1" of Fig.\ref{Fig.example}, when objects in the novel view are occluded in the input views, pixel-based representation focuses on the non-occluded areas, with the occluded parts supplemented by volume-based representation.
\item In ``Case 2" of Fig.\ref{Fig.example}, when objects in the the novel view fall beyond the frustum range for any of the input views, pixel-based representation focuses on the non-truncated areas, with the truncated parts supplemented by volume-based representation.
\item In ``Case 3" of Fig.\ref{Fig.example}, for distant elements out of the volume range, volume-based representation concentrates on reconstruction within the volume, leaving the reconstruction of distant elements to pixel-based representation.
\item In ``Case 4" of Fig.\ref{Fig.example}, for objects with fine-grained details (e.g., cars, lane markings), volume-based representation aims to predict their coarse 3D structures, leaving the surface details to pixel-based representation.
\end{itemize}

\subsection{Additional Comparisons on RealEstate10K}
\label{sec:supl:re10k}
\begin{figure*}[!ht]
\centering
\includegraphics[width=0.85\textwidth]{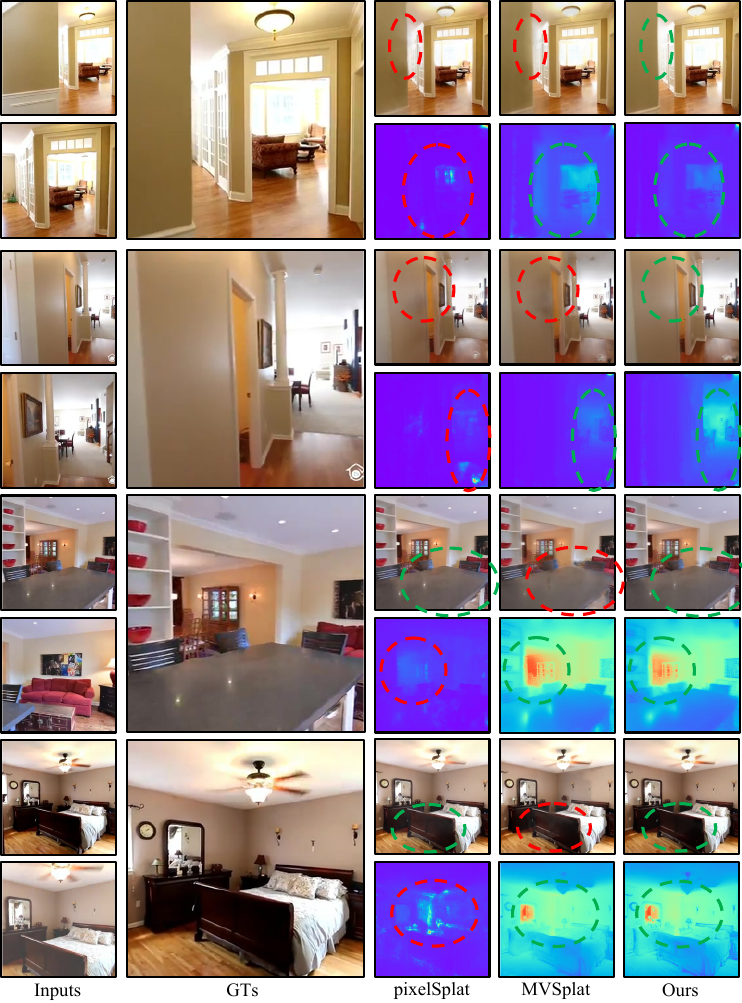}
\caption{Additional qualitative results on scene-centric reconstruction performed on RealEstate10K \cite{re10k2018}. The first two columns are images of input views and ground-truth novel views. The remaining three columns are results generated by pixelSplat \cite{pixelsplat2024}, MVSplat \cite{mvsplat2024} and our method, respectively. The \emph{red dashed lines} highlight undesirable artifacts, while the \emph{green ones} denote plausibly-rendered areas.}
\label{Fig.more_re10k}
\end{figure*}

As shown in Fig.\ref{Fig.more_re10k}, we present more qualitative comparisons with state-of-the-art methods pixelSplat \cite{pixelsplat2024} and MVSplat \cite{mvsplat2024} on RealEstate10K \cite{re10k2018}, a large-scale dataset for scene-centric reconstruction task. We can see that our method can render novel view images and depths with comparable and even superior quality to other methods.

\end{document}